\documentclass[11pt]{article}
\usepackage{latexsym,amsfonts,amsmath,epsfig,amssymb}
\usepackage[colorlinks,linkcolor=red,anchorcolor=blue,urlcolor=red,citecolor=blue]{hyperref}
\usepackage{graphicx,times}
\usepackage{mathrsfs}
\usepackage{caption}
\usepackage{subfigure}
\usepackage{graphicx}
\usepackage{epstopdf}
\usepackage{multirow}
\usepackage{booktabs}
\usepackage{algorithm,algpseudocode}

\usepackage[numbers,sort&compress]{natbib}
\graphicspath{{Figures/}}                 % 定义所有的.eps/.png文件在TTfigures子目录下

\newtheorem{theorem}{Theorem}
\newtheorem{lemma}{Lemma}
\newtheorem{definition}{Definition}
\newtheorem{remark}{Remark}
\newcommand{\R}{\mathbb{R}}
\newcommand{\C}{\mathbb{C}}

\newcommand{\PP}{\mathbb{P}}

\newcommand{\<}{\langle}
\renewcommand{\>}{\rangle}

\newcommand{\vct}[1]{\boldsymbol{#1}}
\newcommand{\mtx}[1]{\boldsymbol{#1}}
\newcommand{\tub}[1]{\mathring{\vct{#1}}}
\newcommand{\tc}[1]{\vec{\vct{#1}}}

\newcommand{\rank}{\operatorname{rank}}
\newcommand{\tr}{\operatorname{Tr}}

\newcommand{\sgn}{\operatorname{sgn}}

\newcommand{\Pg}{{\cal P}_{\Gamma}}
\newcommand{\Pgp}{{\cal P}_{\Gamma'}}
\newcommand{\Pgc}{{\cal P}_{\Gamma^\perp}}
\newcommand{\Pgs}[1]{{\cal P}_{\Gamma_{#1}}}
\newcommand{\PT}{{\cal P}_T}
\newcommand{\PTc}{{\cal P}_{T^\perp}}
\newcommand{\PO}{{\cal P}_{\Omega}}

\newcommand{\Pl}{{\cal P}_{\Lambda}}
\newcommand{\OpId}{\mathcal{I}}

\newcommand{\qed}{\hfill $\Box$}

\newcounter{nextauthor}
\def\mathrm{\mbox}

\numberwithin{remark}{section}

\begin{document}
\title{\Large {\bf Robust Tensor Completion Using Transformed Tensor SVD}}

\author{Guangjing Song\footnotemark[1], \  Michael K. Ng\footnotemark[2], \ and\ Xiongjun Zhang\footnotemark[3]
}

\renewcommand{\thefootnote}{\fnsymbol{footnote}}
\footnotetext[1]{School of Mathematics and Information Sciences, Weifang University,
Weifang 261061, P.R. China (e-mail: sgjshu@163.com).}
\footnotetext[2]{Centre for Mathematical Imaging and Vision and Department of Mathematics, Hong Kong
Baptist University, Kowloon Tong, Hong Kong (e-mail: mng@math.hkbu.edu.hk).
Research supported in part by the HKRGC GRF 12306616, 12200317 and 12300218.}
\footnotetext[3]{School of Mathematics and Statistics and Hubei Key Laboratory of Mathematical Sciences,
Central China Normal University, Wuhan 430079, China (e-mail: xjzhang@mail.ccnu.edu.cn).
Research supported in part by the National Natural Science Foundation of China under grants 11801206, 11571098, 11871025,
Hubei Provincial Natural Science Foundation of China under grant 2018CFB105,
and Fundamental Research Funds for the Central Universities under grant CCNU19ZN017.
}

\renewcommand{\thefootnote}{\arabic{footnote}}

%\date{}
\maketitle \vspace*{0mm}
\begin{center}
\begin{minipage}{5.5in}
$${\bf Abstract}$$
In this paper, we study robust tensor
completion by using transformed tensor singular value decomposition (SVD), which employs unitary transform matrices instead
of discrete Fourier transform matrix that is used in the traditional tensor SVD.
The main motivation is that a lower tubal rank tensor can be obtained by using other unitary
transform matrices than that by using discrete Fourier transform
matrix. This would be more effective for robust tensor completion.
Experimental results for hyperspectral, video and face datasets have shown that the recovery performance
for the robust tensor completion problem
by using transformed tensor SVD
is better in PSNR than that by using Fourier transform and other robust tensor completion methods.
\end{minipage}
\end{center}

\begin{center}
\begin{minipage}{5.5in}

\end{minipage}
\end{center}

\section{Introduction}
Tensor (multi-dimensional arrays) are generalizations of vectors
and matrices, which can be used as a powerful tool in modeling
multi-dimensional data such as videos \cite{liu2013}, color images
\cite{Nguyen2012,konstantions}, hyperspectral images \cite{fan2019hyperspectral, ng2017adaptive, yang2019remote},
and electroencephalography (EEG)
\cite{Cichocki}. Based on its multilinear algebraic properties,
a tensor can take full advantage of its structures to provide better
understanding and higher accuracy of the multi-dimensional data. In
many tensor data applications
\cite{cui2019precondi, Ji2016Tensor, konstantions, Kolda, Miwakeichi, Omberg,Rabanser2017, xie2018kronecker, zhang2018nonconvex}, tensor
data sets are often corrupted and/or incomplete owing to various
unpredictable or unavoidable situations. It is motivated us to
perform tensor completion and tensor robust principal component
analysis for multi-dimensional data processing.

Compared with matrix completion and robust principal component
analysis, tensor completion and tensor robust principal component
analysis are far from being well-studied. The main issues are the
definitions of tensor ranks and tensor decompositions. In the  matrix case,
it has been shown that the nuclear norm is the convex envelope of the matrix
rank over a unit ball of spectral norm \cite{Fazel2002,recht2010guaranteed}.
By solving a convex programming
problem, one can recover a low rank matrix exactly with overwhelming
probability, from a small fraction of its entries, even part of them
are corrupted, provided that the corruptions are reasonably sparse
\cite{Candes2011,Candes2009,chen2013, oymak2015simultaneously, Recht2011}.

Unlike the matrix case, there exist different kinds of definitions of ranks of a tensor.
For instance, the CANDECOMP/PARAFAC (CP) rank is defined as the minimal
number of the rank one outer products of tensors, which is NP-hard
to compute in general \cite{Kolda2009}. Although many authors
\cite{Jain2014,Karlsson2016} have recovered some special low CP rank
tensors by different methods, it is often computationally
intractable to determine the CP rank or its best convex
approximation.
Tensor Train (TT)
rank  \cite{oseledets2011tensor} is generated by the TT
decomposition using the link structure of each core tensor. Since
the link structure, the TT rank is only efficient for higher order
tensor for tensor completion. Bengua et al. \cite{bengua2017efficient}
proposed a novel approach based on TT rank for color images and videos completion.
However, this method may be challenged when the third-dimension of the data is high, such as hyperspectral data.
The Tucker rank (multi-rank) is actually a vector
whose entries can be derived from the factors of Tucker
decomposition \cite{Tucker66}.
Liu et al. \cite{liu2013} proposed to
use the sum of the nuclear norms of unfolding matrices of a tensor to
recover a low Tucker rank tensor.
However, the sum of the nuclear norms of unfolding matrices of a tensor
is not the convex envelope of the sum of ranks of unfolding matrices of a tensor \cite{romera2013new}.
Moreover, Mu et al. \cite{mu2013} showed that the sum of nuclear norms
of unfolding matrices of a tensor is suboptimal and proposed a
square deal method to recover a low rank and high-order tensor.
While the square deal method only utilizes one mode information of unfolding matirces for third-order tensors.
Other extensions can be found in \cite{gandy2011} and references therein. In
\cite{gu2014}, Gu et al. provided a perfect recovery of two
components (the low-rank tensor and the entrywise sparse corruption
tensor) under restricted eigenvalue conditions. In \cite{Huang2014},
Huang et al. proposed a tensor robust principal component analysis
model for exact recovery guarantee under certain tensor incoherence
conditions.

The tensor-tensor product (t-product) and associated algebraic
construction based on the Fourier transform, cosine transform and
any invertible transform for tensors of order three or higher are
studied in \cite{Kilmer2011,kernfel,martin2013}, respectively. With
this framework, Kilmer et al. \cite{Kilmer2011} introduced an
SVD-like factorization called the tensor SVD as well as the
definition of tubal rank. Compared with other tensor decompositions,
this tensor SVD has been shown to be superior in capturing the
spatial-shifting correlation that is ubiquitous in real-world data
\cite{Kilmer2011,martin2013,kilmer,zhang, zhang2019corrected}. Moreover, the tubal
nuclear norm is the convex envelope of the tubal average rank within
the unit ball of the tensor spectral norm. Motivated by the above
results, Zhang et al. \cite{Zhang2017} derived theoretical
performance bounds of the model proposed in \cite{zhang} using the
tensor SVD algebraic framework for third-order tensor recovery from
limited sampling.
Zhou et al. \cite{zhou2018tensor}
proposed a novel factorization method based on the tensor nuclear norm in the Fourier domain
for solving the third-order tensor completion problem.
Hu et al. \cite{hu2017twist} proposed  a twist tensor nuclear norm for tensor completion,
which relaxes the tensor multi-rank of the twist tensor in the Fourier domain.
Being different from tensor completion, robust tensor
completion is more complex due to the sparse noise in the observations.
Jiang and Ng \cite{jiangm} showed that one can
recover a low tubal rank tensor exactly with overwhelming
probability by simply solving a convex program, where the objective
function is a weighted combination of tubal nuclear norm, a convex
surrogate of the tubal-rank, and the $\ell_1$-norm.
Recently, Lu et al. \cite{lu2018tensor}
considered the tensor robust principal component analysis problem
and proposed a tensor nuclear norm based on t-product and tensor SVD in the Fourier domain,
where the theoretical guarantee for the exact
recovery was also provided.

The main aim of this paper is to study robust tensor completion
problems by using transformed tensor SVD,
which employs unitary transform matrices instead
of discrete Fourier transform matrix in the tensor SVD.
The main motivation is that a lower tubal rank tensor can be obtained by using other unitary
transform matrices than that by using discrete Fourier transform
matrix. This would be more effective for robust tensor completion.
The main contributions of this paper are given as follows.
(i) One can
recover a low transformed tubal rank tensor exactly with
overwhelming probability provided that its rank is sufficiently
small and its corrupted entries are reasonably sparse.
Because of the use of unitary transformation,
there are new results in the convex envelope of rank, the subgradient formula and tensor basis required
in the proof.
(ii) We propose a new unitary transformation that can lead to significant recovery results compared with
the use of the Fourier transform.
(iii) Experimental results for hyperspectral and face images and video data have shown that the recovery performance
by using transformed tensor SVD is better in PSNR than that by using Fourier transform in tensor SVD and other tensor completion methods.

The outline of this paper is given as follows. In Section
\ref{sec:pre},  we introduce transformed tensor SVD.
In Section \ref{sec:Theorey}, we analyze the robust tensor
completion problem and the algorithm for solving the model.
In Section \ref{NE}, numerical
results are presented to show that the effectiveness of the proposed tensor
SVD for the robust tensor completion problem.
Finally, some concluding remarks are given in Section \ref{Con}.
All proofs are deferred to the Appendix.

\subsection{Notation and Preliminaries}

Throughout this paper, the fields of real number and complex number
are denoted as $\mathbb{R}$ and $\mathbb{C}$, respectively. Tensors and matrices
are denoted by Euler letters and boldface capital
letters, respectively. For a third-order tensor $\mathcal{A} \in
\mathbb{R}^{n_{1}\times n_{2}\times n_{3}}$, we denote its
$(i,j,k)$-th entry as $\mathcal{A}_{ijk}$ and use the Matlab
notation $\mathcal{A}(i,:,:), \mathcal{A}(:,i,:)$ and
$\mathcal{A}(:,:,i)$ to denote the $i$-th horizontal, lateral and
frontal slices, respectively. Specifically, the frontal slice
$\mathcal{A}(:,:,i)$ is denoted compactly as $\mathcal{A}^{(i)}$.
$\mathcal{A}(i,j,:)$ denotes a tubal fiber
obtained by fixing the first two indices and varying the third index.
Moreover, a tensor tube of size $1\times 1\times n_{3}$ is denoted
as $\tub{a}$ and a tensor column of size $n_1\times1\times n_{3}$ is
denoted as $\overrightarrow{\bf{a}}$.

The inner product of ${\bf A},{\bf B}\in\C^{n_1 \times n_2}$ is given by $\< {\bf A}, {\bf
B}\> = \tr( {\bf A}^{H} {\bf B})$, where ${\bf A}^{H}$ denotes the
conjugate transpose of ${\bf A}$ and $\tr(\cdot)$ denotes the matrix
trace. For a
vector $v\in \mathbb{C}^{n}$, the $l_{2}$-norm is
$\|v\|_{2}=\sqrt{\sum_{i}|v_{i}|^{2}}$. The spectral norm of a
matrix ${\bf A} \in \mathbb{C}^{n_{1}\times n_{2}}$ is denoted as
$\|{\bf A}\|=\max_{i}\sigma_{i}({\bf A})$, where $\sigma_{i}({\bf
A})$ is the $i$-th largest singular value of ${\bf A}$. The nuclear
norm of a matrix is defined as $\| {\bf A} \|_{\ast}=\sum_{i}\sigma_{i}({\bf A})$. For a
tensor $\mathcal{A}$,
the $\ell_1$-norm is defined as $\|\mathcal{A}\|_1 =
\sum_{i,j,k}|\mathcal{A}_{ijk}|$, the infinity norm is defined as
$\|\mathcal{A}\|_{\infty} = \max_{i,j,k}|\mathcal{A}_{ijk}|$ and
the Frobenius norm is defined as $\|\mathcal{A}\|_F =
\sqrt{\sum_{i,j,k}|\mathcal{A}_{ijk}|^2}$. Suppose that ${\tt L}$
is a tensor operator, then its operator norm is defined as
%\begin{equation*}
$\|{\tt L}\|_{\textup{op}} = \sup_{\|\mathcal{A}\|_F \leq 1} \|{\tt
L}(\mathcal{A})\|_F$.

\section{Transformed Tensor Singular Value Decomposition}\label{sec:pre}

Let ${\bf \Phi}$ be the unitary transform matrix with
${\bf \Phi} {\bf \Phi}^H = {\bf \Phi}^H {\bf \Phi} = {\bf I}$, where ${\bf I}$ is the identity matrix.
$\hat{\mathcal{A}}_{{\bf \Phi}}$ represents a
third-order tensor obtained via multiplying by ${\bf \Phi}$ on all tubes
along the third dimension of $\mathcal{A}$, i.e.,
$$
\textrm{vec}(\hat{\mathcal{A}}_{{\bf \Phi}}(i,j,:))={{\bf
\Phi}} \left(\textrm{vec}(\mathcal{A}(i,j,:))\right),
$$
where $\textrm{vec}(\cdot)$ is the vectorization operator that maps
the tensor tube to a vector. Here we write $\hat{\mathcal{A}}_{{\bf
\Phi}}={\bf \Phi}[\mathcal{A}]$. Moreover, one can get $\mathcal{A}$
from $\hat{\mathcal{A}}_{{\bf \Phi}}$  by using ${\bf \Phi}^{H}$ operation
along the third-dimension of $\hat{\mathcal{A}}_{{\bf \Phi}}$,
i.e., ${\cal A} = {\bf\Phi}^H[\hat{\mathcal{A}}_{{\bf \Phi}}]$.

We construct a block
diagonal matrix based on the frontal slices of ${\cal A}$ as follows:
$$
\overline{\mathcal{\mathcal{A}}}=\textrm{blockdiag}(\mathcal{\mathcal{A}}):=
\left(
                                  \begin{array}{cccc}
                                    \mathcal{A}^{(1)} &  &  &  \\
                                     & \mathcal{A}^{(2)}&  & \\
                                     & & \ddots &  \\
                                     &  &  &\mathcal{A}^{(n_{3})}\\
                                  \end{array}
                                  \right),
$$
%where $\mathcal{A}^{(i)}$  is the $i$-th frontal
%slices of ${\cal A},~i=1,\ldots,n_{3}$.
Also
we can convert the block diagonal matrix
into a tensor by the following fold operator:
$$
\textrm{fold}(\textrm{blockdiag}(\mathcal{\mathcal{A}}))=\textrm{
fold}( \overline{\mathcal{\mathcal{A}}} ) := \mathcal{\mathcal{A}}.
%\quad {\rm and} \quad
%\textrm{fold}(\textrm{blockdiag}(\hat{\mathcal{A}}_{\bf \Phi} ))=
%\textrm{fold}( \overline{\mathcal{\mathcal{A}}}_{{\bf \Phi}} ) =
%\hat{\mathcal{A}}_{\bf \Phi}
$$

Kernfeld et al. \cite{kernfel} defined
the $\star_{{\bf L}}$-product between two tensors by the slices
products in the transform domain, where ${\bf L}$ is an arbitrary
invertible transform.
%The definitions of the identity tensor, the
% unitary tensor,  as well as their associated algebraic
%properties, are also considered and studied.
%To facilitate the
%discussion of robust tensor completion problems in Section
%\ref{sec:Theorey},
In this paper, we are mainly interested in the t-product which
is based on unitary transforms.
%Moreover, based on the
%$``\textrm{blockdiag}"$ and $``\textrm{fold}"$ operations, we can
%modify the expression of the  t-product as follows.

\begin{definition}
\label{def1} The ${\bf \Phi}$-product of $\mathcal{A} \in \C^{n_1
\times n_2 \times n_3}$ and $\mathcal{B} \in \C^{n_2 \times n_4
\times n_3}$ is a tensor $\mathcal{C} \in \C^{n_1 \times n_4 \times
n_3}$, which is given by
\begin{equation*}
\mathcal{C} =\mathcal{A} \diamond_{\bf \Phi} \mathcal{B}={\bf
\Phi}^{H}\left[ \textrm{fold} \left(
 \textrm{blockdiag}(\mathcal{\hat{A}}_{\bf \Phi}) \times
\textrm{blockdiag}(\mathcal{\hat{B}}_{\bf \Phi})\right) \right ],
\end{equation*}
where $``\times"$ denotes the standard matrix product.
\end{definition}

%We remark that when the t-product \cite{Kilmer2011} is written as
%Definition \ref{def1}, the associativity rule is valid and the
%relationship between the t-product  and  ${\bf \Phi}$-product is
%straightforward to derive. For example, t
The t-product \cite{Kilmer2011}  of
$\mathcal{A} \in \R^{n_1 \times n_2 \times n_3}$ and $\mathcal{B}
\in \R^{n_2 \times n_4 \times n_3}$  is a tensor $\mathcal{C} \in
\R^{n_1 \times n_4 \times n_3}$ given by
\begin{equation} \label{cir}
\mathcal{C}=\mathcal{A} \ast \mathcal{B}=\textrm{Fold}_{vec} \left
( \textrm{Circ}(\mathcal{A}) \times \textrm{Vec}( \mathcal{B} )
\right ),
\end{equation}
where $\textrm{Fold}_{vec}$ is an operation that takes
$\textrm{Vec}(\mathcal{B})$ into a tensor, i.e.,
$\textrm{Fold}_{vec} ( \textrm{Vec}(\mathcal{B}))=\mathcal{B}$,
$$
\quad \textrm{Vec}(\mathcal{B})=\left(
                       \begin{array}{c}
                         \mathcal{B}^{(1)} \\
                        \mathcal{B}^{(2)} \\
                         \vdots \\
                         \mathcal{B}^{(n_{3})} \\
                       \end{array}
                     \right),
$$
and
$$\textrm{Circ}(\mathcal{A})=\left(
                      \begin{array}{ccccc}
                        \mathcal{A}^{(1)} &  \mathcal{A}^{(n_{3})} &  \mathcal{A}^{(n_{3}-1)} & \cdots &  \mathcal{A}^{(2)} \\
                       \mathcal{A}^{(2)} &  \mathcal{A}^{(1)} & \mathcal{A}^{(n_{3})} & \cdots &  \mathcal{A}^{(3)} \\
                        \vdots &\ddots  & \ddots & \ddots & \vdots \\
                         \mathcal{A}^{(n_{3})} &  \mathcal{A}^{(n_{3}-1)} & \cdots &  \mathcal{A}^{(2)} &  \mathcal{A}^{(1)}\\
                      \end{array}
                    \right).
$$
The t-product (\ref{cir}) can be seen as a special case of Definition \ref{def1}.
Recall that the block circulant matrix $\textrm{Circ}(\mathcal{A})$
can be diagonalized by the discrete Fourier transform matrix
${\bf F}_{n_3}$ and the diagonal matrices are the frontal slices of
$\hat{\mathcal{A}}_{{\bf F}_{n_3}}$, i.e.,
$$
({\bf F}_{n_{3}}\otimes I_{n_{1}}) \times \textrm{Circ}(\mathcal{A})
\times ({\bf F}_{n_{3}}^H \otimes I_{n_{2}}) =
\textrm{blockdiag}(\hat{\mathcal{\mathcal{A}}}_{{\bf F}_{n_3}}),
$$
where $\otimes$ is the Kronecker product. It follows that
\begin{align*}
 \mathcal{A} \ast \mathcal{B}
 &=\textrm{Fold}_{vec} \left ( \textrm{Circ}(\mathcal{A}) \times \textrm{Vec}( \mathcal{B}) \right ) \\
&= \textrm{Fold}_{vec} \Big( ({\bf F}_{n_{3}}^H \otimes I_{n_{1}})
\times \textrm{blockdiag}(\hat{\mathcal{\mathcal{A}}}_{{\bf F}})\times ({\bf F}_{n_{3}}\otimes I_{n_{2}}) \times \textrm{Vec}(\mathcal{B}) \Big) \\
&=\textrm{ Fold}_{vec} \left (  ({\bf F}_{n_{3}}^H \otimes
I_{n_{1}}) \times
\textrm{blockdiag}(\hat{\mathcal{\mathcal{A}}}_{{\bf F}}) \times
\textrm{Vec}(\hat{\mathcal{\mathcal{B}}}_{{\bf F}})\right ) \\
&= \textrm{fold}\left( ({\bf F}_{n_{3}}^H \otimes I_{n_{1}}) \times
\textrm{blockdiag}(\hat{\mathcal{\mathcal{A}}}_{{\bf F}}) \times
\textrm{blockdiag}(\hat{\mathcal{\mathcal{B}}}_{{\bf F}})
\right ) \\
&={\bf F}_{n_{3}}^H \left [ \textrm{fold} \left (
\textrm{blockdiag}(\hat{\mathcal{\mathcal{A}}}_{{\bf
F}})\times \textrm{blockdiag}(\hat{\mathcal{\mathcal{B}}}_{{\bf F}}) \right ) \right ] \\
& =\mathcal{A} \diamond_{{\bf F}_{n_3}} \mathcal{B}.
\end{align*}

According to ${\bf \Phi}$-product, we have the definitions of the conjugate transpose of $\mathcal{A}$, the identity tensor,
the unitary tensor, and the inner product between two tensors.

\begin{definition}
\label{def2.5} The conjugate transpose of $\mathcal{A}\in
\C^{n_{1}\times n_{2}\times n_{3}}$ with respect to ${\bf \Phi}$ is
the tensor $\mathcal{A}^{H}\in \C^{n_{2}\times n_{1}\times n_{3}}$
obtained by
\begin{equation*}
\mathcal{A}^{H} = {\bf \Phi}^{H} \left [ \textrm{fold} \left
(\textrm{blockdiag}(\mathcal{\hat{A}}_{\bf \Phi})^{H}\right
)\right].
\end{equation*}
\end{definition}

\begin{definition} \cite[Proposition 4.1]{kernfel}\label{idd}
The identity tensor $\mathcal{I}_{\bf \Phi} \in \C^{n \times n
\times n_3}$ (with respect to ${\bf \Phi}$) is defined to be a
tensor such that ${\cal I}_{\bf \Phi} = {\bf \Phi}^H [ {\cal T}]$,
where ${\cal T} \in \R^{n \times n \times n_3}$ with each frontal
slice  being the $n \times
n$ identity matrix.
\end{definition}

\begin{definition}  \cite[Definition 5.1]{kernfel}
A tensor $\mathcal{Q} \in \C^{n \times n \times n_3}$ is
%transformed
unitary with respect to ${\bf \Phi}$-product
if it satisfies
\begin{equation*}
\mathcal{Q}^{H} \diamond_{{\bf \Phi}} \mathcal{Q} = \mathcal{Q}
\diamond_{\bf \Phi} \mathcal{Q}^{H} =\mathcal{I}_{\bf \Phi},
\label{eq7}
\end{equation*}
where $\mathcal{I}_{\bf \Phi}$ is the identity tensor.
\end{definition}\label{def5}

\begin{definition}\label{def2.8}
The inner product of $\mathcal{A},\mathcal{B}\in\C^{n_1
\times n_2 \times n_3}$ is defined as
\begin{equation} \label{tprod}
\langle\mathcal{A}, \mathcal{B} \rangle = \sum_{i=1}^{n_{3}}\langle
\mathcal{A}^{(i)}, \mathcal{B}^{(i)}  \rangle =\langle
\overline{\mathcal{A}}_{\bf \Phi}, \overline{\mathcal{B}}_{\bf \Phi}
\rangle.
\end{equation}
\end{definition}

In the above definition, $\langle
\mathcal{A}^{(i)}, \mathcal{B}^{(i)}  \rangle$ is the standard inner product of two matrices.
In addition, a tensor $\mathcal{A}$ is called to be diagonal if each frontal
slice $\mathcal{A}^{(i)}$ is a diagonal matrix \cite{Kilmer2011}.
Based on the above definitions, we have the following
transformed tensor SVD with respect to ${\bf \Phi}$.

\begin{theorem} \cite[Theorem 5.1]{kernfel}\label{them1}
Suppose that $\mathcal{A}\in \C^{n_{1}\times n_{2}\times n_{3}}$.
Then $\mathcal{A}$ can be factorized as follows:
\begin{equation} \label{equ12}
\mathcal{A}= \mathcal{U} \diamond_{\bf \Phi} \mathcal{S}
\diamond_{\bf \Phi} \mathcal{V}^{H},
\end{equation}
where $\mathcal{U}\in \C^{n_{1}\times n_{1}\times n_{3}},$
$\mathcal{V}\in \C^{n_{2}\times n_{2}\times n_{3}}$ are unitary
tensors with respect to ${\bf \Phi}$-product, and $\mathcal{S}\in
\C^{n_{1}\times n_{2}\times n_{3}}$ is a diagonal tensor.
\end{theorem}

The tensors $\mathcal{U}$, $\mathcal{V}$ and $\mathcal{S}$ in the
transformed tensor SVD can be computed by SVDs of $\hat{\mathcal{A}}^{(i)}_{\bf \Phi}$,
which is summarized in
Algorithm \ref{alg1}.

\begin{algorithm}
\caption{Transformed tensor SVD for third-order tensors \cite{kernfel}} \label{alg1}
\textbf{Input:} $\mathcal{A}\in \C^{n_1 \times n_2 \times n_3}$. \\
~~1: $\mathcal{\hat{A}}={\bf \Phi}[\mathcal{A}]$;\\
~~2: \textbf{for} $i=1,...,n_{3}$ \textbf{do}\\
~~3: $[\textbf{U},\textbf{S},\textbf{V}]=\mbox{SVD}(\mathcal{\hat{A}}^{(i)})$;\\
~~4: $\mathcal{\hat{U}}^{(i)}=\textbf{U},~\mathcal{\hat{S}}^{(i)}=\textbf{S},~\mathcal{\hat{V}}^{(i)}=\textbf{V};$\\
~~5:  \textbf{end for}\\
~~6:
 $\mathcal{U}={\bf \Phi}^{H}[\mathcal{\hat{U}}]$, $\mathcal{S}={\bf \Phi}^{H}[\mathcal{\hat{S}}]$,
 $\mathcal{V}={\bf \Phi}^{H}[\mathcal{\hat{V}}]$. \\
\textbf{Output:}  $\mathcal{U}\in \C^{n_1 \times n_1 \times
n_3},~\mathcal{S}\in \C^{n_1 \times n_2 \times n_3},~\mathcal{V}\in
\C^{n_2 \times n_2 \times n_3}.$
\end{algorithm}

\begin{remark}
For computational improvement, we also use the skinny transformed tensor SVD.
For any $\mathcal{A}\in \C^{n_{1}\times n_{2}\times n_{3}}$ with $\emph{rank}(\mathcal{A})=r$ (see in the following definition),
the skinny transformed tensor SVD is given by $\mathcal{A}= \mathcal{U} \diamond_{\bf \Phi} \mathcal{S}
\diamond_{\bf \Phi} \mathcal{V}^{H}$, where $\mathcal{U}\in \C^{n_{1}\times r\times n_{3}},$
$\mathcal{V}\in \C^{n_{2}\times r\times n_{3}}$ are unitary
tensors with respect to ${\bf \Phi}$-product, and $\mathcal{S}\in
\C^{r\times r\times n_{3}}$ is a diagonal tensor.
\end{remark}

%\subsection{Transformed Tubal Tensor Rank and Nuclear Norm}

Based on the transformed tensor SVD given in Theorem \ref{them1}, the tensor tubal rank can be defined as follows.
\begin{definition}
%[Transformed tubal multi-rank and tubal rank]
The tubal multi-rank of a tensor $\mathcal{A} \in
\C^{n_1 \times n_2 \times n_3}$ is a vector $\vct{r} \in \R^{n_3}$
with its $i$-th entry being the rank of the $i$-th frontal slice of
$\hat{\mathcal{A}}_{\bf \Phi}$, i.e., $r_i =
\rank(\hat{\mathcal{A}}^{(i)}_{\bf \Phi})$. The tensor
tubal rank, denoted as $\rank(\mathcal{A})$, is defined as the
number of nonzero singular tubes of $\mathcal{S}$, where
$\mathcal{S}$ comes from the transformed tensor SVD of $\mathcal{A}=\mathcal{U}
\diamond_{\bf \Phi}\mathcal{S}\diamond_{\bf \Phi}\mathcal{V}^{H}$,
i.e.,
\begin{equation}
\rank(\mathcal{A}) = \#\{i: \mathcal{S}(i, i, :) \neq \vct{0}\}
= \max_{i} r_i. \label{eq9}
\end{equation}
\end{definition}

It follows from \cite{jiangm} that the tensor spectral norm of
$\mathcal{A} \in \C^{n_1 \times n_2 \times n_3}$ relate to ${\bf
\Phi}$ , denoted as $\|\mathcal{A}\|$, can be defined as
$\|\mathcal{A}\| = \|\overline{{\bf \Phi}[\mathcal{A}]}\|$. In other
words, the tensor spectral norm of $\mathcal{A}$ equals to the
matrix spectral norm of its block diagonal form in the transform
domain. Moreover, suppose that a tensor operator ${\tt L}$ can be
represented as a tensor $\mathcal{L}$ via $\Phi$-product with
$\mathcal{A}$, we have $\| {\tt L}\|_{\textup{op}} = \|\mathcal
{L}\|$. The aim of this paper is to recover a low transformed
tubal rank tensor, which motivates us to introduce the following definition
of tensor nuclear norm.

\begin{definition}
The transformed tubal nuclear norm of a tensor $\mathcal{A} \in
\C^{n_1 \times n_2 \times n_3}$, denoted as
$\|\mathcal{A}\|_{\textup{TTNN}}$, is the sum of the nuclear norms of
all the frontal slices of $\hat{\mathcal{A}}_{\bf \Phi}$, i.e.,
$\|\mathcal{A}\|_{\textup{TTNN}} =
 \sum_{i=1}^{n_3}
\|{\hat{\mathcal{A}}}^{(i)}_{\bf \Phi}\|_{\ast}$. \label{def9}
\end{definition}

Next we show that
 %Moreover, we can prove that
 the transformed tubal nuclear norm
(TTNN) of a tensor is the convex envelope of the sum of the
elements of the tensor tubal multi-rank over a
unit ball of the tensor spectral norm.
This is why the TTNN is useful for low transformed tubal rank tensor recovery.
We remark this is the new result in the literature, and the proof is different from
%The proof can be found in Appendix.
\cite[Theorem 1]{Fazel2002} because we consider the complex-valued matrices and tensors.

\begin{lemma}\label{lemm1}
For any tensor $\mathcal{X} \in \C^{n_1 \times n_2 \times n_3}$,
$\rank_{sum}(\mathcal{X})=\sum_{i=1}^{n_3}
\rank(\hat{\mathcal{X}}^{(i)}_{\bf \Phi})$ denotes a transformed
tubal multi-rank function. Then $\|\mathcal{X}\|_{\textup{TTNN}}$ is
the convex envelope of the function $\rank_{sum}(\mathcal{X})$ on
the set $\{ \mathcal{X} \ | \ \| \mathcal{X}\| \leq 1\}$.
\end{lemma}

The proof can be found in Appendix A.
%We remark that the proof is different from
%We remark that the main trick used in the proof of
%Lemma \ref{lemm1} follows the lines of
%\cite[Theorem 1]{Fazel2002} because we consider the complex-valued matrices and tensors.
Next we will introduce two kinds of tensor basis
which play  important roles in
tensor coordinate decomposition as well as introducing the tensor
incoherence conditions in the sequel.

\begin{definition} \label{defn}
%[Transformed tensor  basis]
(i) The transformed column basis with respect to ${\bf \Phi}$, denoted as
$\tc{e}_i$, is a tensor of size $n_1 \times 1 \times n_3$ with the
$i$-th tube of ${\bf \Phi} [\tc{e}_i]$ is equal to $\tc{1}$
(each entry in the $i$-th tube is 1) and the rest equaling to 0.
Its associated conjugate transpose $\tc{e}_i^H$ is called
transformed row basis with respect to ${\bf \Phi}$.
(ii) The transformed tube
basis with respect to ${\bf \Phi}$, denoted as $\tub{e}_k$, is a tensor of
size $1 \times 1 \times n_3$ with the $(1,1,k)$-th entry of ${\bf
\Phi}[\tub{e}_k]$ equaling to 1 and the remaining entries equaling to 0.
\label{def10}
\end{definition}

% Here, since  $\tc{e}_i$ and $\tub{e}_k$ are determined
%by different unitary transformations, we don't know their specific
%form.  However, for special unitary transformation, their exact
%expressions can be derived.
%For example, when ${\bf \Phi}$ is chosen
%as the discrete Fourier transform, $\tc{e}_i$ and $\tub{e}_k$ can be
%expressed as Definition 2.12 in \cite{Zhang2017}.

Denote
$\mathcal{E}_{ijk}$  as a unit tensor with the $(i,j,k)$-th
entry equaling to 1 and others equaling to 0. Based on Definition \ref{defn},
 $\mathcal{E}_{ijk}$ can be expressed as
\begin{equation}\label{EE}
\mathcal{E}_{ijk} ={\bf \Phi}[ \tc{e}_i\diamond_{\bf \Phi}
\tub{e}_k \diamond_{\bf \Phi} \tc{e}_j].
\end{equation}
Then for a third-order tensor
$\mathcal{A} \in
\C^{n_1 \times n_2 \times n_3}$, we can decompose it as
\begin{equation}\label{AA}
\mathcal{A} = \sum_{i,j,k} \<\mathcal{E}_{ijk},
\mathcal{A}\>\mathcal{E}_{ijk} = \sum_{i,j,k}
\mathcal{A}_{ijk}\mathcal{E}_{ijk}.
\end{equation}
These properties will be
used many times in the proof of our main results in Section
\ref{sec:Theorey}. These new bases are different from existing bases and they are
required in the proof of unitary transform-based tensor recovery.

\section{Recovery Results by Transformed Tensor SVD} \label{sec:Theorey}

Suppose that we are given a third-order tensor $\mathcal{L}_0$ that has
a low transformed tubal rank with respect to ${\bf \Phi}$
and is also corrupted by a sparse tensor
$\mathcal{E}_0$,
%Here, both $\mathcal{L}_0$ and $\mathcal{E}_0$ are
%of arbitrary magnitude.
where the transformed tubal rank of
$\mathcal{L}_0$ is not known. Moreover,
%Furthermore,
we have no idea about the locations of the nonzero entries of
$\mathcal{E}_0$, not even how many there are.
%In this section, we
%study robust tensor completion problems by using transformed tensor
%singular value decomposition, i.e.,
We would like to recover $\mathcal{L}_0$ from
a set of observed entries of $\mathcal{X}$. We use the TTNN of a
tensor to get a low rank solution and $\ell_1$ norm to get a sparse
solution. Mathematically, the model can be stated as follows:
\begin{equation}
\min_{\mathcal{L},\,\mathcal{E}} \ \|\mathcal{L}\|_{\textup{TTNN}} +
\lambda\|\mathcal{E}\|_{1},\,\,\, \textup{s.t.},\,
\mathcal{P}_\Omega(\mathcal{L} + \mathcal{E}) =
\mathcal{P}_\Omega(\mathcal{X}), \label{mainp}
\end{equation}
where $\lambda$ is a penalty parameter and $\PO$ is a linear
projection such that the entries in the set $\Omega$ are given while
the remaining entries are missing.

We remark that the convex optimization problems constructed in
\cite{zhang,Zhang2017,jiangm} can be seen as special cases of (\ref{mainp}),
which aim to solve the tensor
completion and tensor robust principal component analysis,
respectively. For
instance, if the unitary transform ${\bf \Phi}$ is based on discrete
Fourier transform, the transformed tubal nuclear norm can be
replaced by the tubal nuclear norm (TNN).
%, which means that (\ref{mainp})
%reduces to the main model in \cite{jiangm}.

% It is well known that, if  most entries of
%$\mathcal{L}_0$ are equal to zero, we cannot recovery it exactly.
%Suppose that $\mathcal{L}_0$ is both low-rank and sparse, e.g.,
%$\mathcal{L}_0=\mathcal{E}_{111}$
% \big($(\mathcal{L}_0)_{ijk}=1$ when $i =j =k= 1$ and zeros everywhere else\big), then
%we are not able to identify the low-rank tensor $\mathcal{L}_0$ in
%these cases. To make the problem meaningful,
Here we need some
incoherence conditions on $\mathcal{L}_0$ to ensure that it is not
sparse.

\begin{definition}\label{def3.2}
Assume that $\rank(\mathcal{L}_0) = r$ and its
skinny transformed tensor SVD is $\mathcal{L}_0 = \mathcal{U} \diamond_{\bf \Phi}
\mathcal{S} \diamond_{\bf \Phi}  \mathcal{V}^{H}$. $\mathcal{L}_0$
is said to satisfy the transformed tensor incoherence conditions
with parameter $\mu
> 0$ if
\begin{align}
\max_{i=1, \dots, n_1} \|\mathcal{U}^{H} \diamond_{\bf \Phi} \tc{e}_i\|_F \leq \sqrt{\frac{\mu r}{n_1}}, \label{eq16}\\
\max_{j=1, \dots, n_2} \|\mathcal{V}^{H} \diamond_{\bf \Phi}
\tc{e}_j\|_F\leq \sqrt{\frac{\mu r}{n_2}}, \label{eq17}
\end{align}
and \begin{equation} \|\mathcal{U} \diamond_{\bf \Phi}
\mathcal{V}^{H} \|_{\infty} \leq \sqrt{\frac{\mu r}{n_1 n_2 n_3}},
\label{eq18}
\end{equation}
\label{def13} where $\tc{e}_i$ and $\tc{e}_j$ are the tensor basis
with respect to ${\bf \Phi}$.
\end{definition}

For convenience, we
denote $n_{(1)}=\max(n_{1},n_{2})$ and $n_{(2)}=\min(n_{1},n_{2})$.
The main result of this paper can be stated in the following theorem.

\begin{theorem}\label{TheoremM}
Suppose that $\mathcal{L}_0 \in \C^{n_1 \times n_2 \times n_3}$ obeys
(\ref{eq16})-(\ref{eq18}), and the observation set $\Omega$ is
uniformly distributed among all sets of cardinality $m = \rho n_1
n_2 n_3$. Also suppose that each observed entry is independently
corrupted with probability $\gamma$. Then, there exist universal
constants $c_1, c_2 > 0$ such that with probability at least $1 -
c_1(n_{(1)}n_3)^{-c_2}$, the recovery of $\mathcal{L}_0$ with
$\lambda = 1/\sqrt{\rho n_{(1)}n_3}$ is exact, provided that
\begin{equation}
r \leq \frac{c_rn_{(2)}}{\mu (\log(n_{(1)}n_3))^2}\, \,\,\,
\textup{and} \,\,\, \,\gamma \leq c_\gamma, \label{eq19}
\end{equation}
where $c_r$ and $c_\gamma$ are two positive constants. \label{the1}
\end{theorem}
\begin{remark} By the inner product given in Definition \ref{def2.8}, a direct generalization of the  transformed tensor incoherence conditions listed in \cite{jiangm} for
arbitrary unitary transform are
\begin{align*}
\max_{i=1, \dots, n_1} \|\mathcal{U}^{H} \diamond_{\bf \Phi} \tc{e}_i\|_F \leq \sqrt{\frac{\mu r}{n_{1}n_{3}}}, ~~
\max_{j=1, \dots, n_{2}} \|\mathcal{V}^{H} \diamond_{\bf \Phi}
\tc{e}_j\|_F\leq \sqrt{\frac{\mu r}{n_{2} n_{3}}} ,
\end{align*}
and
\begin{align*} \|\mathcal{U} \diamond_{\bf \Phi}
\mathcal{V}^{H} \|_{\infty} \leq \sqrt{\frac{\mu r}{n_1 n_2 n^2_3}}.
\end{align*}
The right hands of the three inequalities are obviously smaller than those given in \eqref{eq16}-\eqref{eq18},
which means that the exact recovery conditions are weaker than those of \cite{jiangm}.
\end{remark}

The idea of the proof is to employ convex analysis to derive the
conditions in which one can check whether the pair $(\mathcal{L},
\mathcal{E})$ is the unique minimizer to (\ref{mainp}), and
to explicitly show that the conditions in Theorem \ref{the1} are met
with overwhelming probability.
%Our proof follows closely the idea
%presented in \cite{Huang2014,jiangm,xli}, where
The main tools of our proof are
the non-commutative Bernstein Inequality (NBI) and the
golfing scheme \cite{Candes2011,Gross2010}.
The detailed proof is given in Appendix B.

\subsection{Optimization Algorithm}\label{OA}

In this subsection, we develop a symmetric Gauss-Seidel based
multi-block alternating direction method of multipliers (sGS-ADMM)
\cite{Chen2017An, li2016} to solve the robust tensor completion problem
(\ref{mainp}). The sGS-ADMM has been validated the efficiency in
many fields, e.g., see \cite{Chen2017An, li2016, Bai2016, wang2018, zhang2019corrected} and
references therein. Let $\mathcal{L}+\mathcal{E}=\mathcal{M}$.
Problem (\ref{mainp}) can be rewritten as
\begin{equation}\label{ObjF}
\begin{split}
\min_{\mathcal{L},\,\mathcal{E},\, \mathcal{M}} & \
\|\mathcal{L}\|_{\textup{TTNN}} +
\lambda\|\mathcal{E}\|_{1}\\
 \textup{s.t.} & \
\mathcal{L} + \mathcal{E}=\mathcal{M}, \
\mathcal{P}_\Omega(\mathcal{M}) = \mathcal{P}_\Omega(\mathcal{X}).
\end{split}
\end{equation}
The augmented Lagrangian function of (\ref{ObjF}) is defined by
$$
L(\mathcal{L},\,\mathcal{E},\, \mathcal{M},\, \mathcal{Z}):=
\|\mathcal{L}\|_{\textup{TTNN}} + \lambda\|\mathcal{E}\|_{1}-\langle
\mathcal{Z}, \mathcal{L} + \mathcal{E}-\mathcal{M}
\rangle+\frac{\beta}{2}\|\mathcal{L} +
\mathcal{E}-\mathcal{M}\|_F^2,
$$
where $\mathcal{Z}$ is the Lagrangian multiplier and $\beta$ is the
penalty parameter. Let $\mathfrak{D}:=\{\mathcal{M} \in \C^{n_1 \times n_2 \times n_3}:\mathcal{P}_\Omega(\mathcal{M}) = \mathcal{P}_\Omega(\mathcal{X})\}$.
The iteration system of the sGS-ADMM can be
described as follows:
\begin{align}\label{u1211}
&\mathcal{M}^{k+\frac{1}{2}}={\arg\min}_{\mathcal{M}\in\mathfrak{D}}\Big\{L(\mathcal{L}^k,\,\mathcal{E}^k,\, \mathcal{M},\, \mathcal{Z}^k)\Big\},\\\label{w1}
&\mathcal{L}^{k+1}=\arg\min_\mathcal{L}\Big\{L(\mathcal{L},\,\mathcal{E}^k,\, \mathcal{M}^{k+\frac{1}{2}},\, \mathcal{Z}^k)\Big\},\\ \label{u1}
&\mathcal{M}^{k+1}={\arg\min}_{\mathcal{M}\in\mathfrak{D}}\Big\{L(\mathcal{L}^{k+1},\,\mathcal{E}^k,\, \mathcal{M},\, \mathcal{Z}^k)\Big\},\\ \label{z1}
&\mathcal{E}^{k+1}=\arg\min_\mathcal{E}\Big\{L(\mathcal{L}^{k+1},\,\mathcal{E},\, \mathcal{M}^{k+1},\, \mathcal{Z}^k)\Big\},\\ \label{x1}
&\mathcal{Z}^{k+1}=\mathcal{Z}^k-\tau\beta(\mathcal{L}^{k+1} + \mathcal{E}^{k+1}-\mathcal{M}^{k+1}),
\end{align}
where $\tau\in(0,(1+\sqrt{5})/2)$ is the step-length.

The solution with respect to $\mathcal{M}$ can be given by
\begin{equation}
\mathcal{M}= \left\{
\begin{array}{cl}
\mathcal{X}_{ijk}, & \mbox{if}\ (i,j,k)\in\Omega, \\
\Big(\mathcal{L}+\mathcal{E}-\frac{1}{\beta}\mathcal{Z}\Big)_{ijk}, &\mbox{otherwise}.
\end{array}
\right.
\end{equation}

Similar to the proximal mapping of the nuclear norm of a matrix,
we give the proximal mapping of the TTNN of a tensor.
The proximal mapping of $\|\cdot\|_{\textup{TTNN}}$ at $\mathcal{Y}$ can be given in the following theorem.
\begin{theorem}\label{PMTN}
For any $\mathcal{Y}=\mathcal{U} \diamond_{\bf \Phi} \mathcal{S}
\diamond_{\bf \Phi} \mathcal{V}^{H}$,
the minimizer of the following problem
\begin{equation}\label{TNMU}
\min_{\mathcal{X}}\Big\{\lambda\|\mathcal{X}\|_{\textup{TTNN}}+\frac{1}{2}\|\mathcal{X}-\mathcal{Y}\|_F^2\Big\}
\end{equation}
is given by
\begin{equation}\label{TTNNMS}
\emph{Prox}_{\lambda\|\cdot\|_{\textup{TTNN}}}(\mathcal{Y}):= \mathcal{U}\diamond_{\bf \Phi}  \mathcal{S}_{\lambda}
\diamond_{\bf \Phi}  \mathcal{V}^{H},
\end{equation}
where $\mathcal{S}_{\lambda}= \mathbf{\Phi}^H[\hat{\mathcal{S}}_{\lambda}]$
and $\hat{\mathcal{S}}_{\lambda}=\max\{\hat{\mathcal{S}}_{\bf \Phi}-\lambda,0\}$.
\end{theorem}

%\begin{IEEEproof}
By the definition of the TTNN, problem (\ref{TNMU}) can be rewritten as
\begin{equation}\label{UDTNM}
\min_{\mathcal{X}}\Big\{\sum_{i=1}^{n_3}
\lambda\|\hat{\mathbf{X}}_{\bf \Phi}^{(i)}\|_{*}
+\frac{1}{2}\|\hat{\mathbf{X}}_{\bf \Phi}^{(i)}-\hat{\mathbf{Y}}_{\bf \Phi}^{(i)}\|_F^2\Big\}
\end{equation}
By \cite[Theorem 2.1]{Cai2010}, the minimizer of (\ref{UDTNM}) is given by
$$
\hat{\mathbf{X}}_{\bf \Phi}^{(i)}=\hat{\mathbf{U}}_{\bf \Phi}^{(i)}\hat{\mathbf{\Sigma}}_\lambda (\hat{\mathbf{V}}_{\bf \Phi}^{(i)})^H,
$$
where $\hat{\mathbf{\Sigma}}_{\lambda} = \max\{\hat{\mathbf{S}}_{\bf \Phi}^{(i)}-\lambda,0\}$.
By using the inverse unitary transform along the third-dimension, we get that
the optimal solution of (\ref{TNMU}) is given by (\ref{TTNNMS}).
%This completes the proof.
%\end{IEEEproof}

The subproblem with respect to $\mathcal{L}$ in (\ref{w1}) can be
described as
\begin{equation}\label{LOr}
\min \
\|\mathcal{L}\|_{\textup{TTNN}}+\frac{\beta}{2}\Big\|\mathcal{L}
-\Big(\mathcal{M}^{k+\frac{1}{2}}+\frac{1}{\beta}\mathcal{Z}^k-\mathcal{E}^{k}\Big)\Big\|_F^2.
\end{equation}
By Theorem \ref{PMTN}, the minimizer of problem (\ref{LOr}) is given by
\begin{equation}\label{LK}
\mathcal{L}^{k+1}=\mathcal{U} \diamond_{\bf \Phi}
\mathcal{S}_{\beta}\diamond_{\bf \Phi} \mathcal{V}^{H},
\end{equation}
where $\mathcal{M}^{k+\frac{1}{2}}+\frac{1}{\beta}\mathcal{Z}^k-\mathcal{E}^{k}=\mathcal{U} \diamond_{\bf \Phi}
\mathcal{S}\diamond_{\bf \Phi} \mathcal{V}^{H}$ and
$\mathcal{S}_{\beta}={\bf\Phi}^H[\hat{\mathcal{S}}_{\beta}]$ with $\hat{\mathcal{S}}_{\beta}
=\max\{\hat{\mathcal{S}}_{\bf \Phi}-\frac{1}{\beta},0\}$.
The minimizer with respect to $\mathcal{E}$ in (\ref{z1}) can be
given by
\begin{equation}\label{EK}
\mathcal{E}^{k+1}=\mbox{sgn}(\mathcal{H})
\circ\max\Big\{|\mathcal{H}|-\frac{\lambda}{\beta},0\Big\},
\end{equation}
where $\mathcal{H}:=\mathcal{M}^{k+1}+\frac{1}{\beta}\mathcal{Z}^k-\mathcal{L}^{k+1}$,
$\circ$ denotes the point-wise product,
and $\mbox{sgn}(\cdot)$ denotes the signum function.
i.e.,
$$
\mbox{sgn}(y):= \left\{\begin{array}{cl}
1, & \mbox{if} \ y>0,\\
0, & \mbox{if} \ y=0,\\
-1, & \mbox{if} \  y<0.\\
\end{array}\right.
$$

Since only two blocks of the objective function in (\ref{ObjF}) are
nonsmooth and the other block is not involved in (\ref{ObjF}), the sGS-ADMM is convergent
\cite[Theorem 3]{li2016}. The sGS-ADMM for solving (\ref{ObjF}) can
be stated in Algorithm \ref{alg2}.

\begin{algorithm}[h]
\caption{A symmetric Gauss-Seidel based multi-block ADMM for solving
(\ref{ObjF})} \label{alg2}
\textbf{Input:} $\tau\in(0,(1+\sqrt{5})/2), \beta>0,
\mathcal{L}^0, \mathcal{E}^0, \mathcal{Z}^0$. For $k=0,1,2,\ldots,$ perform the following steps: \\
~~1: Compute $\mathcal{M}^{k+\frac{1}{2}}$ by\\
$$
\mathcal{M}^{k+\frac{1}{2}}= \left\{
\begin{array}{cl}
\mathcal{X}_{ijk}, & \mbox{if}\ (i,j,k)\in\Omega, \\
\Big(\mathcal{L}^k+\mathcal{E}^k-\frac{1}{\beta}\mathcal{Z}^k\Big)_{ijk}, &\mbox{otherwise}.
\end{array}
\right.
$$ \\
~~2: Compute $\mathcal{L}^{k+1}$ via (\ref{LK}). \\
~~3: Compute $\mathcal{M}^{k+1}$ by \\
$$
\mathcal{M}^{k+1}= \left\{
\begin{array}{cl}
\mathcal{X}_{ijk}, & \mbox{if}\ (i,j,k)\in\Omega, \\
\Big(\mathcal{L}^{k+1}+\mathcal{E}^k-\frac{1}{\beta}\mathcal{Z}^k\Big)_{ijk}, &\mbox{otherwise}.
\end{array}
\right.
$$
\\
~~4: Compute $\mathcal{E}^{k+1}$ via (\ref{EK}).\\
~~5: Update $\mathcal{Z}^{k+1}$ by (\ref{x1}).  \\
~~6: If a stopping criterion is not met, set $k:=k+1$ and go to step
1.
\end{algorithm}

\section{Experimental Results}\label{NE}

In this section, numerical results are presented to show the
effectiveness of the proposed method for robust tensor completion.
We compare the transformed tensor SVD with
the sum of nuclear norms of unfolding matrices of a tensor
plus a sparse tensor (SNN)\footnote{\footnotesize https://tonyzqin.wordpress.com/} \cite{Goldfarb2014},
tensor SVD using Fourier transform (t-SVD (Fourier))\footnote{\footnotesize https://canyilu.github.io/publications/} \cite{lu2018tensor},
and low-rank tensor completion by parallel matrix factorization
(TMac)\footnote{\footnotesize https://xu-yangyang.github.io/TMac/} \cite{xu2013parallel}.
All the experiments are performed under Windows 7 and MATLAB R2018a
running on a desktop (Intel Core i7, @ 3.40GHz, 8.00G RAM).
%SNN\footnote{\footnotesize https://tonyzqin.wordpress.com/} \cite{Goldfarb2014}
%t-SVD\footnote{\footnotesize https://canyilu.github.io/publications/} \cite{Lu2016, lu2018tensor},
%TMac \cite{xu2013parallel}
%Kronecker-basis-representation (KBR)\footnote{\footnotesize http://gr.xjtu.edu.cn/web/dymeng/3} \cite{xie2018kronecker}

\subsection{Experimental Setting}

The sampling ratio of observations is defined as $\rho := \frac{|\Omega|}{n_1n_2n_3}$,
where $\Omega$ is generated uniformly at random and $|\Omega|$
denotes the number of the entries of $\Omega$.
%The relative error is
%defined by
%$$
%\mbox{Rel}:=\frac{\|\mathcal{L}-\mathcal{L}_0\|_F}{\|\mathcal{L}\|_F},
%$$
%where $\mathcal{L}$ is the recovered solution and $\mathcal{L}_0$ is
%the ground-truth tensor.
In order to evaluate the performance of different methods for real-world tensors,
the peak signal-to-noise ratio (PSNR) is used to
measure the quality of the estimated tensors, which is defined as follows:
$$
\mbox{PSNR}:=10\log_{10}
\frac{n_1n_2n_3({\mathcal{L}}_{\max}-{\mathcal{L}}_{\min})^2}{\|\mathcal{L}-{\mathcal{L}}_0\|_F^2},
$$
where $\mathcal{L}$ is the recovered solution, $\mathcal{L}_0$ is
the ground-truth tensor, ${\mathcal{L}}_{\max}$ and ${\mathcal{L}}_{\min}$
are maximal and minimal entries of $\mathcal{L}_0$, respectively.

As suggested by Theorem \ref{TheoremM},
we set $\lambda=\frac{a}{\sqrt{n_{(1)}n_3}}$ and adjust it slightly to obtain the best possible results.
In all experiments, $a$ is selected from $\{1.1, 1.3, 1.7, 1.8, 2\}$ in Fourier transform
and from $\{10, 15, 18, 20, 23, 25, 28, 30, 33, 35, 40, 45, 50\}$ in unitary and wavelet transforms.
%we choose
%$\lambda$ based on $1/\sqrt{\rho n_{(1)}n_3}$ in our experiments
%and adjust it slightly around this range to obtain the best possible results.
Moreover,
$\tau$ is set to be $1.618$ for its convergence \cite{li2016} and $\beta$ is chosen from
$\{0.01,0.05,0.1,0.5\}$ to get the highest PSNR values in Algorithm \ref{alg2}.
%\ref{alg2}.
The Karush-Kuhn-Tucker (KKT) conditions of problem (\ref{ObjF}) are given by
$$
\left\{\begin{array}{lll}
\mathcal{Z}\in\partial \|\mathcal{L}\|_{\textup{TTNN}}, \  \mathcal{Z}\in\partial\lambda\|\mathcal{E}\|_1,  \\
\mathcal{L} + \mathcal{E}=\mathcal{M}, \
\mathcal{P}_\Omega(\mathcal{M}) = \mathcal{P}_\Omega(\mathcal{X}),
\end{array}\right.
$$
where $\partial \|\mathcal{L}\|_{\textup{TTNN}}$ and $\partial\lambda\|\mathcal{E}\|_1$  denote the subdifferentials of $ \|\cdot\|_{\textup{TTNN}}$ and $\lambda\|\cdot\|_1$, respectively.
Note that $\mathcal{P}_\Omega(\mathcal{M}) = \mathcal{P}_\Omega(\mathcal{X})$
is always satisfied in each iteration of the sGS-ADMM.
We measure the accuracy of an approximate optimal solution by using
the following relative KKT residual:
$$
\eta_{res}:=\max\{\eta_z,\eta_e,\eta_m\},
$$
where
\[
\begin{split}
&
\eta_z=\frac{\|\mathcal{L}-\mbox{Prox}_{\|\cdot\|_{\textup{TTNN}}}(\mathcal{Z}+\mathcal{L})\|_F}{1+\|\mathcal{Z}\|_F+\|\mathcal{L}\|_F}, \eta_e = \frac{\|\mathcal{E}-\mbox{Prox}_{\lambda\|\cdot\|_1}(\mathcal{Z}+\mathcal{E})\|_F}{1+\|\mathcal{Z}\|_F+\|\mathcal{E}\|_F},\\
& \eta_m =
\frac{\|\mathcal{L}+\mathcal{E}-\mathcal{M}\|_F}{1+\|\mathcal{L}\|_F+\|\mathcal{E}\|_F+\|\mathcal{M}\|_F}.
\end{split}
\]
Here $\mbox{Prox}_g$ is the proximal mapping of $g$, i.e.,
$\mbox{Prox}_g(y) = \arg\min_x\{g(x)+\frac{1}{2}\|x-y\|^2\}$. The
stopping criterion of the Algorithm %\ref{alg2}
is set to $\eta_{res}\leq
5\times10^{-4}$ and the maximum number of iterations is set to be 500.

For the sparse level of $\mathcal{E}$, a fraction $\gamma$
of its entries are uniformly corrupted by additive independent and identically distributed
noise from a standard Gaussian distribution $N(0, 1)$ at random,
which generates the sparse tensor $\mathcal{E}$.
The testing real-world tensors are rescaled in $[0,1]$.

\begin{figure}[!t]
\centering
\subfigure{
\begin{minipage}[b]{0.99\textwidth}
\centerline{\scriptsize } \vspace{1.5pt}
              \includegraphics[width=6.1in,height=3.9in]{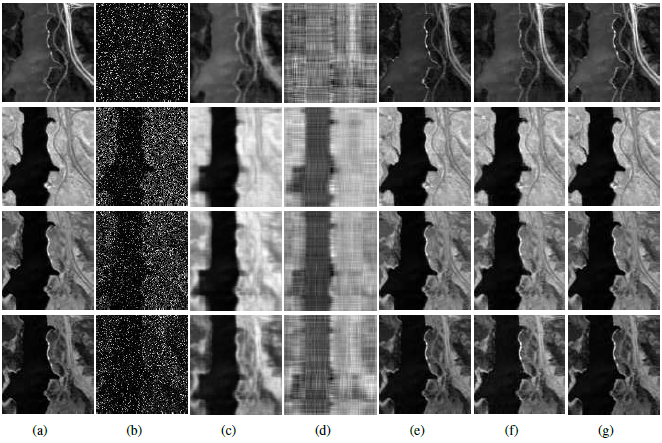}
\end{minipage}
}
       \caption{\small
       (a) Original images for Japser Ridge dataset;
       (b) Observed images ($60\%$ sampling ratio and $30\%$ corrupted entries);
       (c) Recovered images by SNN [PSNR = 26.00];
       (d) Recovered images by TMac [PSNR = 16.47];
       (e) Recovered images by t-SVD (Fourier) [PSNR = 33.63];
       (f) Recovered images by t-SVD (wavelet) [PSNR = 33.59];
       (g) Recovered images by t-SVD (data) [PSNR = 37.38].} \label{HP20603}
\end{figure}

\begin{table}[!t]
\scriptsize
  \begin{center}
          \setlength{\abovecaptionskip}{-1pt}
       \setlength{\belowcaptionskip}{-1pt}
   \caption{\small PSNR values obtained by SNN, TMac, t-SVD (Fourier), t-SVD (wavelet), and t-SVD (data)
   for hyperspectral datasets. The boldface number is the best and the underline number is the second best.}\label{HPRPCA}
  \medskip
\begin{tabular}{|c|  c | c |c c c c c |} \hline
\multirow{2}{*}{} & \multirow{2}{*}{$\rho$}
& \multirow{2}{*}{$\gamma$}&  \multirow{2}{*}{SNN} &  \multirow{2}{*}{TMac}&  t-SVD  & t-SVD & t-SVD   \\
&   &  &     &     & (Fourier)        & (wavelet) & (data)    \\ \hline \hline
\multirow{6}{*}{Samson} & \multirow{3}{*}{$0.6$ }  & 0.1&30.22&23.15& 38.53  & \underline{45.43} &{\bf 53.30} \\
                        &                          & 0.2&29.63&19.35& 34.80  & \underline{41.29} &{\bf 50.68} \\
                        &                          & 0.3&28.06&16.82& 32.26  & \underline{38.22} &{\bf 45.87}\\ \cline{2-8}
                        & \multirow{3}{*}{$0.8$ }  & 0.1&32.43&23.28& 40.87  & \underline{47.34} &{\bf 54.40}\\
                        &                          & 0.2&30.84&19.42& 36.82  & \underline{44.69} &{\bf 52.49} \\
                        &                          & 0.3&29.35&16.86& 33.58  & \underline{39.46} &{\bf 48.28}\\ \hline
\multirow{6}{*}{Japser Ridge} & \multirow{3}{*}{$0.6$ }  & 0.1& 30.13 &21.73& 39.22  & \underline{40.60} & {\bf 45.13}  \\
                        &                          & 0.2& 27.92 &18.76& 36.38  & \underline{37.20} & {\bf 41.13} \\
                        &                          & 0.3& 26.00 &16.47& \underline{33.63}  & 33.59 & {\bf 37.38} \\ \cline{2-8}
                        & \multirow{3}{*}{$0.8$ }  & 0.1& 31.98 &21.84& 40.78  & \underline{42.64} & {\bf 46.76} \\
                        &                          & 0.2& 29.61 &18.81& 37.88  & \underline{38.87} & {\bf 43.13} \\
                        &                          & 0.3& 27.49 &16.51& 35.15  & \underline{35.81} & {\bf 39.33} \\ \hline
\multirow{6}{*}{Urban} & \multirow{3}{*}{$0.6$ }  & 0.1& 27.88&22.20& 38.78   &\underline{39.10}  & {\bf 47.76} \\
                        &                          & 0.2& 26.13&18.43& \underline{36.08}   & 35.70  & {\bf 44.51} \\
                        &                          & 0.3 & 24.69&16.06& \underline{33.39}   & 32.16  & {\bf 39.63}\\ \cline{2-8}
                        & \multirow{3}{*}{$0.8$ }  & 0.1& 30.31&22.29& 40.76   &\underline{42.94} & {\bf 49.76} \\
                        &                          & 0.2& 27.89&18.45& 37.77   &\underline{38.44} & {\bf 45.98} \\
                        &                          & 0.3& 26.06&16.07&\underline{34.98}   & 33.37 & {\bf 42.26}\\ \hline
    \end{tabular}
  \end{center}
\end{table}

\begin{figure}[!t]
\centering
\subfigure{
\begin{minipage}[b]{0.99\textwidth}
\centerline{\scriptsize } \vspace{1.5pt}
              \includegraphics[width=6.1in,height=3.2in]{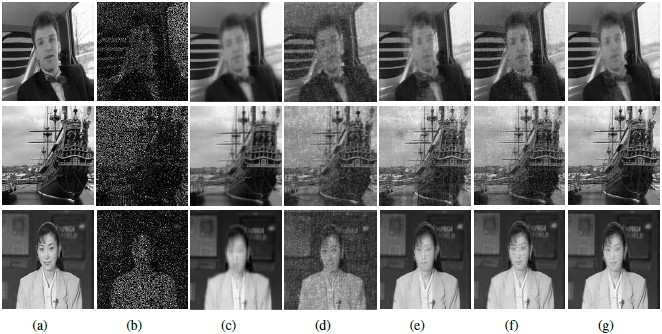}
\end{minipage}
}
       \caption{\small Recovered images by SNN, TMac, t-SVD (Fourier), t-SVD (wavelet) and t-SVD (data) in robust tensor completion
       for video data with $60\%$ sampling ratio and $20\%$ corrupted entries.
        (a) Original images.
       (b) Observed images.
       (c) SNN.
       (d) TMac.
       (e) t-SVD (Fourier).
       (f) t-SVD (wavelet).
       (g) t-SVD (data). }\label{VVCDC}
\end{figure}

\begin{table}[!t]
\scriptsize
  \begin{center}
          \setlength{\abovecaptionskip}{-1pt}
       \setlength{\belowcaptionskip}{-1pt}
   \caption{\small The PSNR values of video data by SNN, TMac, t-SVD (Fourier), t-SVD (wavelet),
   and t-SVD (data).
   The boldface number is the best and the underline number is the second best.}\label{RPCAVD}
  \medskip
\begin{tabular}{|c|  c | c |c c c c c | } \hline

& \multirow{2}{*}{$\rho$} & \multirow{2}{*}{$\gamma$} &  \multirow{2}{*}{SNN} &  \multirow{2}{*}{TMac}& t-SVD & t-SVD   & t-SVD  \\
& &   &      &     & (Fourier)     & (wavelet)& (data) \\ \hline \hline
\multirow{6}{*}{Carphone}    & \multirow{3}{*}{$0.6$} & 0.1 &26.80&20.86& 30.70  & \underline{31.21}  &{\bf 32.38} \\
                           &                        & 0.2 &24.88&17.35& \underline{28.82}  & 28.30  &{\bf 30.14}\\
                           &                        & 0.3 &23.39&15.19& 27.21  & \underline{27.32}  &{\bf 28.09} \\  \cline{2-8}
                           & \multirow{3}{*}{$0.8$ }& 0.1 &28.74&21.03& 32.57  & \underline{32.73}  &{\bf 34.06}\\
                           &                        & 0.2 &26.35&17.47& 30.13  & \underline{30.25}  &{\bf 31.29} \\
                           &                        & 0.3 &24.68&15.27& 28.16  & \underline{28.18}  &{\bf 29.10}  \\   \hline
\multirow{6}{*}{Galleon}    & \multirow{3}{*}{$0.6$} & 0.1 &24.56&20.47& 27.44 & \underline{29.18}  & {\bf 29.80} \\
                           &                        & 0.2&22.14&17.19& 25.63 & \underline{26.55}   & {\bf 27.55} \\
                           &                        & 0.3&19.07&15.12& 24.05 & \underline{24.26}   & {\bf 25.84}  \\  \cline{2-8}
                           & \multirow{3}{*}{$0.8$} & 0.1 &26.86&20.75& 29.37 & \underline{30.67}   & {\bf 31.71}  \\
                           &                        & 0.2 &24.32&17.37& 27.01 & \underline{28.44}   & {\bf 29.03}  \\
                           &                        & 0.3 &22.10&15.23& 25.08 & \underline{26.26}   & {\bf 26.93} \\   \hline
\multirow{6}{*}{Announcer}    & \multirow{3}{*}{$0.6$} & 0.1 &29.58&21.14& 37.90  & \underline{38.24} & {\bf 39.44} \\
                           &                        & 0.2 &27.55&17.52& \underline{35.35}  & 35.07& {\bf 36.57} \\
                           &                        & 0.3 &26.36&15.33& \underline{33.02}  & 31.94& {\bf 34.14} \\  \cline{2-8}
                           & \multirow{3}{*}{$0.8$} & 0.1  &31.57&21.24& 39.62  &\underline{40.45} & {\bf 41.28}  \\
                           &                        & 0.2  &28.67&17.58& \underline{36.61}  & 35.91& {\bf 37.97}  \\
                           &                        & 0.3&27.50&15.37& \underline{34.15}  & 33.54& {\bf 35.23}   \\   \hline
    \end{tabular}
  \end{center}
\end{table}

In the following test, we consider two different kinds of transformations in the proposed method.
The first one is a Daubechies 4 (db4)
discrete wavelet transform in the periodic mode \cite{Daubechies1992ten}
to compute transformed tensor SVD (called t-SVD (wavelet)).
The second one is based on data to construct a unitary transform matrix.
We note that $\mathcal{A}$ is unfolded into a matrix $\mathbf{A}$ along the third-dimension (called t-SVD (data)).
Then we take the singular value decomposition of the unfolding matrix $\mathbf{A}=\mathbf{U}\mathbf{\Sigma} \mathbf{V}^H$.
It is interesting to see that ${\bf U}^H$ is the optimal transformation to obtain a low rank matrix of ${\bf A}$:
$$
\min_{ {\rm rank}({\bf B})=k, \ {\rm unitary} \ {\bf \Phi} } \ \| {\bf \Phi} {\bf A} - {\bf B} \|_F^2.
$$
In practice, the initial estimator $\mathcal{A}$ in the robust tensor completion problem by using the Fourier transform
(i.e., t-SVD completion method) can be used to generate ${\bf \Phi}$.

\subsection{Hyperspectral Data}

In this subsection, we use three hyperspectral datasets: Samson, Japser Ridge, and Urban datasets \cite{zhu2014spectral}
to show the effectiveness of the proposed method.
% by using unitary transform instead of using Fourier and wavelet transforms.
The testing datasets are  third-order tensors (length $\times$ width $\times$ channels).
We describe the three datasets in the following:
\begin{itemize}
\item For the Samson dataset,
we only utilize a region of $95\times 95$ in each image,
where each pixel is recorded at
156 frequency channels covering the wavelengths from 401$nm$ to 889$nm$. Then the spectral
resolution is highly up to 3.13$nm$.
Thus, the size of the resulting tensor is $95\times 95\times 156$.
\item For the Japser Ridge dataset, each pixel is recorded at 224 frequency channels
with wavelengths being from 380$nm$ to 2500$nm$.
 The spectral resolution is up to 9.46$nm$.
 Since this hyperspectral image is too complex to get the ground truth,
 a subimage of $100 \times 100$  pixels is considered.
 The first pixel starts from the $(105,269)$th pixel in the original image.
Due to dense water vapor and atmospheric effects, we only remain 198 channels.
Therefore, the size of the resulting tensor is $100\times 100\times 198$.
\item For the Urban dataset, there are $307 \times 307$ pixels of each image,
each of which corresponds to a $2 \times 2$ $m^2$ area. In this image,
there are 210 wavelengths ranging from 400$nm$  to 2500$nm$, which results in a spectral resolution of 10$nm$.
162 channels of this dataset is remained due to dense water vapor and atmospheric effects.
Hence, the size of the resulting tensor is $307\times 307\times 162$.
\end{itemize}

We consider robust tensor completion problem
for the testing hyperspectral data with different sampling ratios and $\gamma$.
Figure \ref{HP20603} displays the visual comparisons of different methods
for the Japser Ridge dataset with $60\%$ sampling ratio and $30\%$ corruption entries.
We can observe that the visual quality obtained by t-SVD (data) is better
than that obtained by SNN, TMac, t-SVD (Fourier), and t-SVD (wavelet).
The PSNR values obtained by different methods are displayed in Table \ref{HPRPCA}.
We can observe that the PSNR values obtained by t-SVD (data)
are much higher than those obtained by SNN, TMac, t-SVD, and t-SVD (wavelet)
for different sampling ratios $(0.6, 0.8)$ and $\gamma$ $(0.1, 0.2, 0.3)$.
The improvements of t-SVD (data) are very impressive, especially for small $\gamma$.
The performance of t-SVD (wavelet) is better than that of SNN, TMac, and t-SVD (Fourier)
in terms of PSNR values for the Samson and Japser Ridge datasets.
For the Urban dataset, the PSNR values obtained by t-SVD (Fourier) are slightly higher
than those by t-SVD (wavelet), especially for large $\gamma$.

\subsection{Video Data}

In this subsection, we present three video data (length $\times$ width $\times$ frames) including
Carphone ($144\times 176\times 180$), Galleon ($144\times 176\times 200$), and
Announcer ($144\times 176\times 200$)\footnote{\footnotesize{https://media.xiph.org/video/derf/}}
to show the effectiveness of the proposed method in robust tensor completion problem,
where the first channels of all frames in the original data are used.
We just choose $180$ and $200$ frames for these videos to improve the computational time.

We display the visual comparisons of the testing data
 in robust tensor compeltion
with $60\%$ sampling ratio and $20\%$ corruption entries
by SNN, TMac, t-SVD (Fourier), t-SVD (wavelet), and t-SVD (data)
in Figure \ref{VVCDC}.
We can see that the images recovered by t-SVD (data)
are better than those recovered by SNN, TMac, t-SVD (Fourier), and t-SVD (wavelet) in terms of visual quality.
The t-SVD (data) can keep more details than SNN, TMac, t-SVD (Fourier), and t-SVD (wavelet)
for the three testing videos.

\begin{figure*}[!t]
\centering
\subfigure{
\begin{minipage}[b]{0.99\textwidth}
\centerline{\scriptsize } \vspace{1.5pt}
              \includegraphics[width=6.1in,height=3.2in]{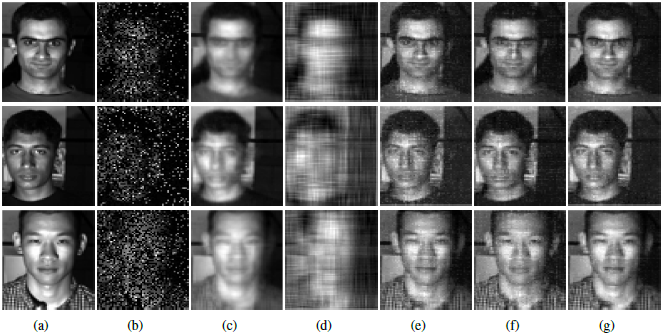}
\end{minipage}
}
       \caption{\small Recovered images by SNN, TMac, t-SVD (Fourier), t-SVD (wavelet),
       and t-SVD (data) in robust tensor completion
       for the extended Yale face database B with $60\%$ sampling ratio and $20\%$ corrupted entries.
       (a) Original images.
       (b) Observed images.
       (c) SNN.
       (d) TMac.
       (e) t-SVD (Fourier).
       (f) t-SVD (wavelet).
       (g) t-SVD (data). }\label{YFDVC}
\end{figure*}

We also show the PSNR values obtained by SNN, TMac, t-SVD (Fourier), t-SVD (wavelet), and t-SVD (data)
for the Carphone, Galleon and Announcer data  with different sampling ratios ($0.6, 0.8$)
and $\gamma$ ($0.1, 0.1, 0.3$) in Table \ref{RPCAVD}.
It can be seen that the PSNR values obtained by  t-SVD (data)
are higher than those by SNN, TMac, t-SVD (Fourier), and t-SVD (wavelet).
The PSNR values obtained by  t-SVD (data) can be improved
around 2dB compared with those by t-SVD (Fourier) for these data.
For the Carphone and Galleon videos, the performance of t-SVD (wavelet)
is better than that of t-SVD (Fourier) in terms of PSNR values.
While the PSNR values obtain by t-SVD (Fourier) is slightly higher
than those obtained by SNN, TMac, and t-SVD (wavelet)
for large $\gamma$ such as $0.2$ and $0.3$ cases.

\subsection{Face Data}

In this subsection, we use the extended Yale face
database B\footnote {http://vision.ucsd.edu/$\sim$iskwak/ExtYaleDatabase/ExtYaleB.html}
to test the robust tensor completion problem.
To improve the computational time,
we crop the original image to contain the face and resize it to $73\times 55$.
Moreover, we only choose first  30 subjects and 25 illuminations in our test.
Then the size of the testing tensor is $73\times 55\times 750$.

%In Fig. \ref{YaleDD}, we display the distributions of singular values
%of all frontal slices of the extended Yale face database B by using different transforms.
%We can observe that the singular values of frontal slices by using wavelet
%and unitary transforms are smaller than those by using Fourier transform.
%In addition, the number of singular values being smaller than $0.01$
%by using wavelet transform is larger than that by using unitary transform.
%However, when the range of singular values is $(0.01,0.1]$,
%the number of singular values of all frontal slices by using unitary
%transform is larger than that by using Fourier and wavelet transforms.
%The  total number of  singular values being smaller than 1 by using wavelet transform
%is larger than that by using unitary transform.

%\begin{figure}[!t]
%  \centering
%       \setlength{\abovecaptionskip}{0pt}
%       \setlength{\belowcaptionskip}{0pt}
%       \includegraphics[width=3.1in,height=1.35in]{YaleD-11.eps}
%       \caption{Distribution of singular values of all frontal slices by using different
%transforms for the extended Yale face database B.}\label{YaleDD}
%\end{figure}%

\begin{table}[!t]
\centering
\scriptsize
  \begin{center}
          \setlength{\abovecaptionskip}{-1pt}
       \setlength{\belowcaptionskip}{-1pt}
   \caption{\small
   The PSNR values of face data by SNN, TMac, t-SVD (Fourier), t-SVD (wavelet),
   and t-SVD (data).
   The boldface number is the best and the underline number is the second best.} \label{YaleRPCA}
  \medskip
  \begin{tabular}{|  c | c |c c c c  c |} \hline
  \multirow{2}{*}{$\rho$} &\multirow{2}{*}{$\gamma$}& \multirow{2}{*}{SNN} & \multirow{2}{*}{TMac}&t-SVD& t-SVD & t-SVD   \\
                              &         &       &      &   (Fourier)        &(wavelet) & (data)  \\   \hline \hline
      \multirow{3}{*}{$0.6$ } & 0.1     & 24.57 &19.51 & 26.00    & \underline{26.85}    & {\bf 28.76}  \\
                              & 0.2     & 22.89 &17.60 & 24.06    & \underline{24.73}    & {\bf 26.26} \\
                              & 0.3     & 21.76 &15.77 & 22.22    & \underline{22.97}    & {\bf 23.75}  \\   \hline
     \multirow{3}{*}{$0.8$ }  & 0.1     & 26.59 &19.54 & 28.07    & \underline{28.86}    & {\bf 30.67}  \\
                              & 0.2     & 24.53 &17.79 & 25.54    & \underline{26.02}    & {\bf 27.60}   \\
                              & 0.3     & 23.06 &15.93 & 23.59    & \underline{24.14}    & {\bf 25.29}  \\   \hline
    \end{tabular}
  \end{center}
\end{table}

The visual comparisons of SNN, TMac, t-SVD (Fourier), t-SVD (wavelet), and t-SVD (data)
for the extended Yale face database B are show
in Figure \ref{YFDVC}, where the sampling ratio is $0.6$ and $\gamma=0.2$.
It can be  seen that the images obtained by t-SVD (data) are more clear
than those obtained by SNN, TMac, t-SVD (Fourier), and t-SVD (wavelet).

In Table \ref{YaleRPCA}, we show the PSNR values of different sampling ratios $(0.6, 0.8)$
and $\gamma$ $(0.1, 0.2, 0.3)$
 for the extended Yale face database B in the robust tensor completion.
It can be seen that the PSNR values obtained
by t-SVD (data) are higher than those obtained
by SNN, TMac, t-SVD (Fourier), and t-SVD (wavelet) for these sampling ratios and $\gamma$.
The PSNR values of t-SVD (data) can be improved around 2dB than those of t-SVD (Fourier).

\section{Concluding Remarks}\label{Con}

We have studied the robust tensor completion problems by using
transformed tensor SVD, which employs other
unitary transform matrices instead of discrete Fourier transform
matrix. The algebraic structure of the associated tensor product
between two tensors is not necessary to be known in general and the
tensor product can be defined via the unitary transform directly.
With this generalized tensor product, we have shown that one can
recover a low transformed tubal rank tensor exactly with
overwhelming probability provided that its rank is sufficiently
small and its corrupted entries are reasonably sparse.
Moreover, we have proposed an ``optimal'' data-dependent transform method in robust tensor completion
problem for third-order tensors. Numerical
examples on many real-word tensors show the usefulness of the transformed tensor SVD method with wavelet and data-dependent
transforms, and demonstrate that the performance of the proposed method is better than that
of existing tensor completion methods.

\section*{Appendix A. Proof of Lemma \ref{lemm1}}\label{Lemma2}

For convenience, we denote
$\Upsilon(\mathcal{X})=\rank_{sum}(\mathcal{X}),$ and $n_{(2)}=\min(n_{1},n_{2})$.
If the spectral
norm of $\mathcal{X}$ is less than or equal to $1$, the conjugate of
transformed  tubal multi-rank function $\Upsilon(\mathcal{X})$ on a
unit ball of the tensor spectral norm can be defined as
\begin{equation*}
\Upsilon ^{\sharp }\left( \mathcal{Y}\right) =\sup_{\left\Vert
\mathcal{X}\right\Vert \leq 1}\left( \textrm{Re}(\langle \mathcal{Y},\mathcal{X}
\rangle) -\rank_{sum}(\mathcal{X}) \right) .
\end{equation*}
Then by the von Neumann's
theorem and the tensor
inner product given in Definition \ref{def2.8}, we can get
\begin{equation}\label{301}
 \textrm{Re}(\langle \mathcal{Y},\mathcal{X} \rangle) = \textrm{Re}(\langle
\mathcal{\hat{Y}}_{\Phi},\mathcal{\hat{X}}_{\Phi} \rangle)
=\sum^{n_{3}}
_{i=1}\textrm{Re}\Big(\tr\Big((\mathcal{\hat{Y}}^{(i)}_{\Phi})^{H}\times
\mathcal{\hat{X}}_{\Phi}^{(i)}\Big)\Big) \leq \sum
_{i=1}^{n_{(2)}n_{3}}\sigma _{i}\left(
\overline{\mathcal{Y}}_{\Phi}\right) \sigma _{i} \left(
\overline{\mathcal{X}}_{\Phi}\right),
\end{equation}
where $\sigma _{i} \left( \overline{\mathcal{X}}_{\Phi}\right) $
denotes the $i$-th largest singular value of
$\overline{\mathcal{X}}_{\Phi}.$ Let
$\mathcal{Y}=\mathcal{U}_{\mathcal{Y}}\diamond_{\Phi}
\Sigma_{\mathcal{Y}}\diamond_{\Phi}\mathcal{V}_{\mathcal{Y}}^{H}$
and $\mathcal{X}=\mathcal{U}_{\mathcal{X}}\diamond_{\Phi}
\Sigma_{\mathcal{X}} \diamond_{\Phi}\mathcal{V}_{\mathcal{X}}^{H}$.
Since $\|\mathcal{X}\|\leq 1$, we can choose
$\mathcal{U}_{\mathcal{X}}=\mathcal{U}_{\mathcal{Y}}$ and
$\mathcal{V}_{\mathcal{X}}=\mathcal{V}_{\mathcal{Y}}$. Thus
\begin{eqnarray*}
\textrm{Re}(\langle \mathcal{Y},\mathcal{X} \rangle)&=&\sum^{n_{3}}
_{i=1}\textrm{Re}\big(\tr((\mathcal{\hat{Y}}^{(i)}_{\Phi})^{H}\times
\mathcal{\hat{X}}_{\Phi}^{(i)} )\big)\\
&=&\textrm{Re}\Big(\tr\left(\overline{\mathcal{V}}_{\mathcal{Y}}\times
\overline{\Sigma}_{\mathcal{Y}}\times
\overline{\mathcal{U}}_{\mathcal{Y}}^{H}\times
\overline{\mathcal{U}}_{\mathcal{Y}}\times \overline{\Sigma}_{\mathcal{X}}\times \overline{\mathcal{V}}_{\mathcal{Y}}^{H}\right) \Big)\\
& = &   \textrm{Re}\Big(\tr \left( \overline{\mathcal{V}}_{\mathcal{Y}}\times
\overline{\Sigma}_{\mathcal{Y}}\times  \overline{\Sigma}_{\mathcal{X}}\times \overline{\mathcal{V}}_{\mathcal{Y}}^{H}\right)\Big) \\
%& = & \tr \left( \overline{\Sigma}_{\mathcal{Y}}\times  \overline{\Sigma}_{\mathcal{X}}\right) \\
&= & \sum _{i=1}^{n_{(2)}n_{3}}\sigma _{i}\left(
\overline{\mathcal{Y}}_{\Phi}\right) \sigma _{i} \left(
\overline{\mathcal{X}}_{\Phi}\right) ,
\end{eqnarray*}%
which shows that the equality in (\ref{301}) can be obtained.
Therefore, the conjugate  of transformed  tubal multi-rank function
can be rewritten as
\begin{equation}\label{SZC}
\Upsilon ^{\sharp }\left( Y\right) =\sup_{\left\Vert\mathcal{
X}\right\Vert \leq 1}\left(\sum _{i=1}^{n_{(2)}n_{3}}\sigma
_{i}\left( \overline{\mathcal{Y}}_{\Phi}\right) \sigma _{i} \left(
\overline{\mathcal{X}}_{\Phi}\right)  -\rank_{sum}(\mathcal{X})
\right).
\end{equation}%
Now we first consider separated cases with the
$\rank_{sum}(\mathcal{X})$ being from $0$ to $n_{(2)}n_{3}$. If
$\rank_{sum}(\mathcal{X})=0$, then $\Upsilon ^{\sharp }\left(
\mathcal{Y}\right) =0$ for all $\mathcal{Y}$. If $
\rank_{sum}(\mathcal{X}) =r,$ then
\begin{align*}
\Upsilon ^{\sharp }\left( \mathcal{Y}\right)=\max \Big\{0,\sigma
_{1}\left( \overline{\mathcal{Y}}_{\Phi}\right)
-1,\ldots,\sum_{i=1}^{r}\sigma _{i}\left(
\overline{\mathcal{Y}}_{\Phi}\right)
-r,\ldots,
\sum_{i=1}^{n_{(2)}n_{3}}\sigma _{i}\left(
\overline{\mathcal{Y}}_{\Phi}\right) -n_{(2)}n_{3}\Big\}.
\end{align*}%
for $1\leq r\leq n_{(2)}n_{3}$.
Furthermore, if $\sigma_{i}(\overline{\mathcal{Y}}_{\Phi})-1>0$, the
right hand of (\ref{SZC}) would be larger. Thus
$$
\Upsilon ^{\sharp }\left( \mathcal{Y}\right) = \left\{
\begin{array}{cl}
0, & \mbox{if}\ \ \left\Vert \overline{\mathcal{Y}}_{\Phi}\right\Vert \leq 1, \\
\sum\limits_{i=1}^{q}  \sigma _{i}\left(
\overline{\mathcal{Y}}_{\Phi}\right) -q,
&\mbox{if}\ \sigma_{q}\left( \overline{\mathcal{Y}}_{\Phi}\right) >1\text{ and }\sigma_{q+1}\left( \overline{\mathcal{Y}}_{\Phi}\right) <1, \ 1\leq q\leq
r.
\end{array}
\right.
$$
The conjugate of $\Upsilon ^{\sharp }\left(\mathcal{Y}\right)$ is
defined as
\begin{align*}
\Upsilon ^{\sharp \sharp} \left( \mathcal{Z}\right)
&=\sup_{\mathcal{Y}}\left( \textrm{Re}(\langle \mathcal{Z},\mathcal{Y} \rangle)
-\Upsilon ^{\sharp }\left( \mathcal{Y}\right)
\right)=\sup_{\overline{\mathcal{Y}}_{\Phi}}\left(\tr\left(
(\overline{\mathcal{Z}}_{\Phi})^{H}\overline{\mathcal{Y}}_{\Phi}\right)
-\Upsilon ^{\sharp }\left( \mathcal{Y}_{\Phi}\right) \right)
\end{align*}%
for all $\mathcal{Z}\in \C^{n_1 \times n_2 \times n_3}$.
In a similar vein, suppose that
$\mathcal{Z}=\mathcal{U}_{\mathcal{Z}}\diamond_{\Phi}
\Sigma_{\mathcal{Z}}\diamond_{\Phi}\mathcal{V}_{\mathcal{Z}}^{H}$
and $\mathcal{Y}=\mathcal{U}_{\mathcal{Y}}\diamond_{\Phi}
\Sigma_{\mathcal{Y}} \diamond_{\Phi}\mathcal{V}_{\mathcal{Y}}^{H}$,
by choosing $\mathcal{U}_{\mathcal{Y}}=\mathcal{U}_{\mathcal{Z}}$ and
$\mathcal{V}_{\mathcal{Y}}=\mathcal{V}_{\mathcal{Z}}$, we obtain that
\begin{equation*}
\Upsilon ^{\sharp \sharp} \left( \mathcal{Z}\right)
=\sup_{\mathcal{Y}}\left( \sum _{i=1}^{n_{(2)}n_{3}}\sigma
_{i}\left(\overline{\mathcal{Z}}_{\Phi}\right) \sigma _{i}\left(
\overline{\mathcal{Y}}_{\Phi}\right) -\Upsilon ^{\sharp }\left(
\mathcal{Y}\right) \right).
\end{equation*}
In the following, we can consider
 two cases, i.e., $\left\Vert
\overline{\mathcal{Z}}_{\Phi}\right\Vert \geq 1$ and $\left\Vert
\overline{\mathcal{Z}}_{\Phi}\right\Vert \leq 1.$
In the first case, if $\left\Vert \overline{\mathcal{Z}}_{\Phi} \right\Vert \geq 1,$
then $\sigma _{1}\left( \overline{\mathcal{Y}}_{\Phi}\right)$ can be
chosen large enough such that
\begin{align*}
\Upsilon ^{\sharp \sharp} \left( \mathcal{Z}\right)
 &
=\sup_{\mathcal{Y}}\left( \sum _{i=1}^{n_{(2)}n_{3}}\sigma
_{i}\left(\overline{\mathcal{Z}}_{\Phi}\right) \sigma _{i}\left(
\overline{\mathcal{Y}}_{\Phi}\right) -\Upsilon ^{\sharp }\left(
\mathcal{Y}\right)
\right)  \\
& =\sup_{\mathcal{Y}}\left( \sum _{i=1}^{n_{(2)}n_{3}}\sigma
_{i}\left(\overline{\mathcal{Z}}_{\Phi}\right) \sigma _{i}\left(
\overline{\mathcal{Y}}_{\Phi}\right)-\left( \sum_{i=1}^{q}\sigma
_{i}\left( \overline{\mathcal{Y}}_{\Phi}\right)
-q\right) \right)  \\
& =\sup_{\mathcal{Y}}\big( \sigma _{1}\left(
\overline{\mathcal{Y}}_{\Phi}\right) \left( \sigma _{1}\left(
\overline{\mathcal{Z}}_{\Phi}\right) -1\right) +\left( \sum
_{i=2}^{n_{(2)}n_{3}}\sigma
_{i}\left(\overline{\mathcal{Z}}_{\Phi}\right) \sigma _{i}\left(
\overline{\mathcal{Y}}_{\Phi}\right) -\left( \sum _{i=2}^{q}\sigma
_{i}\left( \overline{\mathcal{Y}}_{\Phi}\right) -q\right) \right)\Bigg)
\end{align*}%
tends to infinity.
% Then we have $\Upsilon ^{\sharp \sharp} \left( \mathcal{Z}\right) \rightarrow\infty$.
In the second case, $\left\Vert
\overline{\mathcal{Z}}_{\Phi}\right\Vert \leq 1,$ if $\left\Vert
\overline{\mathcal{Y}}_{\Phi}\right\Vert \leq 1$, we obtain that
$\Upsilon ^{\sharp  }\left( \mathcal{Y}\right) =0$
and the supremum is achieved at $%
\sigma _{i}\left( \mathcal{\overline{\mathcal{Y}}}_{\Phi}\right) =1,
i=1,...,n_{(2)}n_{3}$, which yields
$$
\Upsilon ^{\sharp \sharp }\left( \mathcal{Z}\right) =\sum
_{i=1}^{n_{(2)}n_{3}}\sigma
_{i}\left(\overline{\mathcal{Z}}_{\Phi}\right)
=\|\mathcal{Z}\|_{\textup{TTNN}}.
$$
If $\left\Vert \overline{\mathcal{Y}}_{\Phi}\right\Vert >1$, we can
prove
$
\Upsilon ^{\sharp \sharp }\left( \mathcal{Z}\right) \leq
\|\mathcal{Z}\|_{\textup{TTNN}}.
$
In fact,
\begin{align*}
& \sum _{i=1}^{n_{(2)}n_{3}}\sigma
_{i}\left(\overline{\mathcal{Z}}_{\Phi}\right) \sigma _{i}\left(
\overline{\mathcal{Y}}_{\Phi}\right)
-\sum _{i=1}^{q}\left( \sigma _{i}\left( \overline{\mathcal{Y}}_{\Phi}\right)-1\right)  \\
%=&\sum _{i=1}^{n_{(2)}}\sigma
%_{i}\left(\overline{\mathcal{Z}}_{\Phi}\right) \sigma _{i}\left(
%\overline{\mathcal{Y}}_{\Phi}\right)-\sum _{i=1}^{q}\left( \sigma
%_{i}\left( \overline{\mathcal{Y}}_{\Phi}\right) -1\right)-\sum
%_{i=1}^{n_{(2)}n_{3}}\sigma _{i}\left(
%\overline{\mathcal{Z}}_{\Phi}\right)\\
%~~&  +\sum
%_{i=1}^{n_{(2)}n_{3}}\sigma _{i}\left(
%\overline{\mathcal{Z}}_{\Phi}\right)  \\
 =&\sum _{i=1}^{q}\sigma
_{i}\left(\overline{\mathcal{Z}}_{\Phi}\right) \sigma _{i}\left(
\overline{\mathcal{Y}}_{\Phi}\right) +\sum
_{i=q+1}^{n_{(2)}n_{3}}\sigma _{i}\left(
\overline{\mathcal{Z}}_{\Phi}\right) \sigma _{i}\left(
\overline{\mathcal{Y}}_{\Phi}\right)  -\sum _{i=1}^{q}\left( \sigma_{i}
\left( \overline{\mathcal{Y}}_{\Phi}\right) -1\right) -\sum _{i=1}^{n_{(2)}n_{3}}\sigma _{i}\left(
\overline{\mathcal{Z}}_{\Phi}\right)
+\sum
_{i=1}^{n_{(2)}n_{3}}\sigma _{i}\left(
\overline{\mathcal{Z}}_{\Phi}\right)  \\
 =&\sum _{i=1}^{q}\left( \sigma _{i}\left( \overline{\mathcal{Z}}_{\Phi}\right) -1\right) \left(
\sigma _{i}\left( \overline{\mathcal{Y}}_{\Phi}\right) -1\right)
+\sum _{i=q+1}^{n_{(2)}n_{3}}\sigma _{i}\left(
\overline{\mathcal{Z}}_{\Phi}\right) \left( \sigma _{i}\left(
\overline{\mathcal{Y}}_{\Phi}\right) -1\right) +\sum
_{i=1}^{n_{(2)}n_{3}}\sigma _{i}\left( \overline{\mathcal{Z}}_{\Phi}\right)\\
\leq & \sum _{i=1}^{n_{(2)}n_{3}}\sigma _{i}\left( \overline{\mathcal{Z}}_{\Phi}\right),
\end{align*}%
which can be derived from
\begin{equation*}
\sum _{i=1}^{q}\left(
\sigma _{i}\left( \overline{\mathcal{Z}}_{\Phi}\right) -1\right)
\left( \sigma _{i}\left( \overline{\mathcal{Y}}_{\Phi}\right)
-1\right) \leq 0 ~~\text{and}~\sum _{i=q+1}^{n_{(2)}n_{3}}\sigma
_{i}\left( \overline{\mathcal{Z}}_{\Phi}\right) \left( \sigma
_{i}\left( \overline{\mathcal{Y}}_{\Phi}\right) -1\right) \leq 0.
\end{equation*}
In summary, we can get
\begin{equation*}
\Upsilon ^{\sharp \sharp }\left( \mathcal{Z}\right) =\sum
_{i=1}^{n_{(2)}n_{3}}\sigma _{i}\left(
\overline{\mathcal{Z}}_{\Phi}\right) =\left\Vert
\mathcal{Z}\right\Vert_{\textup{TTNN}}
\end{equation*}%
over the set $\left\Vert \mathcal{Z}\right\Vert \leq 1$.

In addition, the convex envelope of
$\rank_{sum}(\mathcal{X})$ on $\left \{ \mathcal{X}\in \C^{n_1 \times
n_2 \times n_3}\left \vert \text{ }\left \Vert {\mathcal{X}}\right
\Vert \leq c \right. \right \}$ can be  changed into
$\rank_{sum}(\mathcal{Y})$ on $\left \{ \mathcal{Y}\in \C^{n_1 \times
n_2 \times n_3}\left \vert \text{ }\left \Vert {\mathcal{Y}} \right
\Vert \leq 1\right.\right \}$ by setting
$\mathcal{Y}=\frac{1}{c}\mathcal{X}$.

\section*{Appendix B. Proof of Theorem \ref{TheoremM}}\label{Them2}

In this section, we provide the detailed proof of Theorem \ref{TheoremM}.
The idea is to employ convex analysis to derive conditions in
 which one can check whether the pair $(\mathcal{L}_0, \mathcal{E}_0)$
 is the unique minizer to (\ref{mainp}), and to show  that
 such conditions are met with overwhelming probability in the
 conditions of Theorem \ref{TheoremM}. Before giving the detailed proof,
 we need to introduce the sampling schemes,  some useful lemmas,
 as well as the subgradients of tensor $l_{1}$ norm and transformed
tubal nuclear norm used in Theorem \ref{TheoremM}, respectively.

\subsection*{Sampling Schemes}\label{sec5:sub1}
The sampling strategy used in Theorem \ref{TheoremM} is the uniform
sampling without replacement. There are other widely used sampling
models, e.g., Bernoulli sampling, adaptive sampling and random
sampling with replacement. To facilitate our proof, we will consider
the independent and  identically distributed \textit{(i.i.d.)
Bernoulli-Rademacher model}. More precisely, we assume $\Omega =
\{(i, j, k) \ | \ \delta_{ijk} = 1\}$, where $\delta_{ijk}'s$ are
i.i.d. Bernoulli variables taking value one with probability $\rho$
and zero with probability $1-\rho$. Such a Bernoulli sampling is
denoted by $\Omega \sim \textup{Ber}(\rho)$ for short. As a proxy
for uniform sampling, the probability of failure under Bernoulli
sampling with $\rho = \frac{m}{n_1n_2n_3}$ closely approximates the
probability of failure under uniform sampling.

Let a subset $\Lambda \subset \Omega$ be the corrupted entries of
$\mathcal{L}_0$ and $\Gamma \subset \Omega$ be locations where data are
available and clean. In a standard Bernoulli model, we suppose that
\begin{displaymath}
\Omega \sim \textup{Ber}(\rho), \,\,\, \Lambda \sim
\textup{Ber}(\gamma\rho), \,\,\, \Gamma \sim \textup{Ber}((1 -
\gamma)\rho),
\end{displaymath}
and the signs of the nonzero entries of $\mathcal{E}_0$ are
deterministic. It has been shown to be much easier to work with a
stronger assumption that the signs of the nonzero entries of
$\mathcal{E}_0$ are independent symmetric $\pm1$ random variables
(i.e., Rademacher random variables). We introduce two independent
random subsets of $\Omega$
\begin{displaymath}
\Lambda' \sim \textup{Ber}(2\gamma\rho), \,\,\, \Gamma' \sim
\textup{Ber}((1 - 2\gamma)\rho),
\end{displaymath}
and it is convenient to consider $\mathcal{E}_0 =
\Pg(\mathcal{E})$ for some fixed tensor $\mathcal{E}$. Assume that a
random sign tensor $\mathcal{M}$ has i.i.d. entries such that for
any index $(i, j, k)$, $\PP(\mathcal{M}_{i,j,k} = 1) =
\PP(\mathcal{M}_{i,j,k} = -1) = \frac{1}{2}$. Then $|\mathcal{E}|\circ
\mathcal{M}$ has components with symmetric random signs. By
introducing a new noise tensor $\mathcal{E}'_0 =
\Pgp(|\mathcal{E}|\circ\mathcal{M})$ and using the standard
derandomization theory \cite[Theorem 2.3]{Candes2011}), we have the following
results.

\begin{lemma}
Suppose that $\mathcal{L}_0$ obeys the conditions of Theorem \ref{TheoremM}
and $\mathcal{E}_0, \mathcal{E}'_0$ are given as above. If
the recovery of $(\mathcal{L}_0, \mathcal{E}'_0)$ is exact with high
probability, it is also exact with at least the same probability for
the model with input data $(\mathcal{L}_0, \mathcal{E}_0)$.
\label{the2}
\end{lemma}

According to Lemma \ref{the2}, we
%Therefore from now on, we
can equivalently assume that the nonzero entries have symmetric
random signs and
\begin{equation}
\Lambda \sim \textup{Ber}(2\gamma\rho), \,\,\, \Gamma \sim
\textup{Ber}((1 - 2\gamma)\rho), \label{eq22}
\end{equation}
for the locations of nonzero and zero entries of $\mathcal{E}_0$,
respectively.

\subsection*{Useful Lemmas}
Suppose that $\mathcal{L}_{0}$ satisfies
\begin{equation}\label{svd1}
\mathcal{L}_{0}=\mathcal{U}\diamond_{\bf
\Phi}\mathcal{S}\diamond_{\bf \Phi}\mathcal{V}^H, \end{equation}
where $\mathcal{U}\in \C^{n_{1}\times r\times n_{3}},$
$\mathcal{V}\in \C^{n_{2}\times r\times n_{3}}$ are unitary tensors,
respectively, and $\mathcal{S}\in \C^{r\times r\times n_{3}}$ is a
diagonal tensor.
 Denote the set $T$ by
\begin{equation}
T = \big\{\mathcal{U}\diamond_{\bf \Phi}\mathcal{Y}^{H} +
\mathcal{W}\diamond_{\bf \Phi} \mathcal{V}^{H} ~|~ \mathcal{Y} \in
\C^{n_2 \times r \times n_3},~\mathcal{W} \in \C^{n_1 \times r
\times n_3} \big\}. \label{e1}
\end{equation}
For any $\mathcal{Z}\in\mathbb{C}^{n_1\times n_2\times n_3}$, the
projections onto $T$ and its complementary are given as
follows \cite[Proposition 7.1]{zhang2019corrected}:
\[
\begin{split}
P_{T}(\mathcal{Z})& = \mathcal{U}\diamond_{\bf \Phi}
\mathcal{U}^{H}\diamond_{\bf \Phi} \mathcal{Z}
+\mathcal{Z}\diamond_{\bf \Phi} \mathcal{V}\diamond_{\bf \Phi}
\mathcal{V}^{H}- \mathcal{U}\diamond_{\bf \Phi}
\mathcal{U}^{H}\diamond_{\bf \Phi} \mathcal{Z}\diamond_{\bf \Phi}
\mathcal{V}\diamond_{\bf \Phi} \mathcal{V}^{H}, \\
P_{T^{\perp}}(\mathcal{Z})& =  (\mathcal{I}_{{\bf \Phi}} -
\mathcal{U}\diamond_{{\bf \Phi}} \mathcal{U}^{H})\diamond_{\bf \Phi}
\mathcal{Z}\diamond_{\bf \Phi} (\mathcal{I}_{{\bf \Phi}} -
\mathcal{V}\diamond_{\bf \Phi}
\mathcal{V}^{H}).
\end{split}
\]

From the definition of $P_{T}(\mathcal{A})$, and recall the inner
product of two tensors, we can
derive the following results.
\begin{lemma} \label{lem6.2}
Suppose that $T$ is defined as $(\ref{e1})$,
then $\langle
P_{T}(\mathcal{A}),\mathcal{B}\rangle=\langle\mathcal{A},P_{T}(\mathcal{B})\rangle.$
\end{lemma}

It follows from Lemma \ref{lem6.2} that $\langle
P_{T}(\mathcal{A}),P_{T^{\perp}}(\mathcal{B})\rangle=0$, where
$\mathcal{A}$ and $\mathcal{B}$ are arbitrary tensors with proper
sizes. Furthermore, based on the transformed tensor incoherence
conditions given in (\ref{eq16})-(\ref{eq17}), we can prove the
following results which will be used many times in the sequel.
\begin{lemma} \label{lem6.3}
Let $\mathcal{A} \in \C^{n_1 \times n_2 \times n_3}$ be an arbitrary
tensor, and $T$ be given as (\ref{e1}). Suppose that the transformed
tensor incoherence conditions (\ref{eq16})-(\ref{eq17}) are
satisfied and denote $\tc{e}_i\diamond _{\bf \Phi}\tub{e}_k\diamond_{\bf \Phi} \tc{e}^{H}_j=(\mathcal{E}_{{\bf \Phi}})_{ijk}$, then
$$\|P_{T}(\tc{e}_i\diamond _{\bf \Phi}\tub{e}_k\diamond_{\bf \Phi} \tc{e}^{H}_j)\|^{2}_{F}=\|P_{T}(\mathcal{E}_{{\bf \Phi}})_{ijk})\|^2_{F}\leq \frac{2\mu_{0}r}{n_{(2)}}.$$
\end{lemma}

%Before moving on,  we need to pay attention to the following
%conclusions which play an important role in proving Lemmas
%\ref{le1}-\ref{le3}.
%Let $\mathcal{E}_{ijk}$ be a unit tensor with
%only $(i,j,k)$-th entry
% equaling to 1 and zeros everywhere else. Then based on Definition \ref{defn}, we can get $\mathcal{E}_{ijk}={\bf
%\Phi}[\tc{e}_i\diamond_{\bf \Phi}\tub{e}_k\diamond_{\bf
%\Phi}\tc{e}_j^{H}]$.
For any given tensor, we have
$\mathcal{Z} \in \C^{n_1 \times n_2 \times n_3}$,
  \begin{align*}
{\bf \Phi}[P_{T}(\mathcal{Z})] &=
 \sum_{i,j,k}\left\langle {\bf \Phi}[P_{T}(\mathcal{Z})],\mathcal{E}_{ijk}\right\rangle\mathcal{E}_{ijk}\\
 &=
 \sum_{i,j,k}\left\langle {\bf \Phi}[P_{T}(\mathcal{Z})],
 {\bf\Phi}[\tc{e}_i\diamond_{\bf \Phi}\tub{e}_k\diamond_{\bf
\Phi}\tc{e}_j^{H}]\right\rangle
 {\bf
\Phi}[\tc{e}_i\diamond_{\bf \Phi}\tub{e}_k\diamond_{\bf
\Phi}\tc{e}_j^{H}] \\\nonumber
 &=\sum_{i,j,k}\left\langle P_{T}(\mathcal{Z}),(\mathcal{E}_{{\bf \Phi}})_{ijk}\right\rangle
 {\bf
\Phi}[(\mathcal{E}_{{\bf \Phi}})_{ijk}],
 \end{align*}
where the last equality follows from
 the definition of tensor inner product.
By applying ${\bf \Phi}^H$ operation to the tubes along the third
dimension of ${\bf \Phi}[P_{T}(\mathcal{Z})]$ and ${\bf
\Phi}[\tc{e}_i\diamond_{\bf \Phi}\tub{e}_k\diamond_{\bf
\Phi}\tc{e}_j^{H}]$, respectively, we get that
\begin{align}
P_{T}(\mathcal{Z}) &= \sum_{i,j,k}\left\langle P_{T}(\mathcal{Z}),(\mathcal{E}_{{\bf \Phi}})_{ijk}\right\rangle(\mathcal{E}_{{\bf \Phi}})_{ijk} =\sum_{i,j,k}\left\langle\mathcal{Z},~P_{T}((\mathcal{E}_{{\bf \Phi}})_{ijk})\right\rangle(\mathcal{E}_{{\bf \Phi}})_{ijk}.\label{j01}
\end{align}

Moreover, setting $\Omega \sim \mbox{Ber}(\rho),$ then we have
\begin{equation}
\begin{split}
 \rho^{-1}P_{T}P_{\Omega}P_{T}(\mathcal{Z})\rho^{-1}\sum_{i,j,k}\delta_{ijk}\left\langle\mathcal{Z},~P_{T}((\mathcal{E}_{{\bf \Phi}})_{ijk})\right\rangle
P_{T}((\mathcal{E}_{{\bf \Phi}})_{ijk}),\label{j02}
\end{split}
\end{equation}
which implies
\begin{equation*}
\begin{split}
 \rho^{-1}\overline{P_{T}P_{\Omega}P_{T}(\mathcal{Z})}
=\rho^{-1}\sum_{i,j,k}\delta_{ijk}\left\langle\mathcal{Z},P_{T}((\mathcal{E}_{{\bf \Phi}})_{ijk})\right\rangle\overline{P_{T}((\mathcal{E}_{{\bf \Phi}})_{ijk})}.
\end{split}
\end{equation*}
Based on the tensor decompostion forms given in
(\ref{j01})-(\ref{j02}), and  applying the methods used in
\cite{jiangm}, we can get  the following three lemmas.
%%which play key roles in proof of our main results.
%The methods used here are similar as those in , so we only list the
%results without their proofs.
\begin{lemma} \label{le1}
Suppose that $\Omega \sim Ber(\rho),$ and $T$ is defined in
(\ref{e1}). Then with high probability,
$$\|\rho^{-1}\mathcal{P}_{T}\mathcal{P}_{\Omega}\mathcal{P}_{T}-\mathcal{P}_{T}\|_{op}\leq \epsilon,$$
provided that $\rho\geq C_0  \frac{\mu r
\log(n_{(1)}n_3)}{n_{(2)}}\epsilon^{-2}$ for some numerical constant $C_0 > 0$.
\end{lemma}
\begin{lemma}\label{le2}
Suppose that $\mathcal{Z} \in \C^{n_1 \times n_2 \times n_3}$ is a fixed
tensor and $\Omega \sim \textup{Ber}(\rho)$. Then with high
probability,
\begin{equation}
\|(\rho^{-1}\PT\PO\PT - \PT)\mathcal{Z}\|_{\infty} \leq \epsilon
\|\mathcal{Z}\|_{\infty}, \label{eq24}
\end{equation}
provided that $\rho \geq C_0 \frac{\mu r
\log(n_{(1)}n_3)}{n_{(2)}}\epsilon^{-2}$ for some numerical constant $C_0 > 0$.
\label{lem2}
\end{lemma}

\begin{lemma}\label{le3}
Suppose that $\mathcal{Z} \in \C^{n_1 \times n_2 \times n_3}$ is a fixed
tensor and $\Omega \sim \textup{Ber}(\rho)$. Then with high
probability,
\begin{equation}
\|(\OpId_{\bf \Phi} - \rho^{-1}\PO)\mathcal{Z}\| \leq
C'_0\sqrt{\frac{n_{(1)}n_3\log(n_{(1)}n_3)}{\rho}}\|\mathcal{Z}\|_{\infty},
\label{eq25}
\end{equation}
provided that $\rho \geq C_0  \frac{\log(n_{(1)}n_3)}{n_{(2)}n_3}$
for some numerical constants $C_0, C'_0 > 0$. \label{lem3}
\end{lemma}

The following lemma is similar to \cite[Lemma 4.5]{Lu2016}
%(the
%difference lists in the definition of the tensor spectral norm),
which provides a upper bound for the spectral norm (which is based on any unitary transform instead of FFT) of the tensors
consisting of Bernoulli sign variables.  For simplicity, we
omit its proof.
\begin{lemma}
For the $n_1 \times n_2 \times n_3$ Bernoulli sign tensor
$\mathcal{M}$ whose entries are distributed as
\begin{equation}
\mathcal{M}_{ijk} = \left\{
\begin{array}{lll}
1, & \textup{with \ probability}\,\,\frac{\rho}{2},\\
0, & \textup{with \ probability}\,\,1-\rho,\\
-1, & \textup{with \ probability}\,\,\frac{\rho}{2},\\
\end{array}
\right. \label{eq26}
\end{equation}
there exists a function $\varphi(\rho)$ satisfying
$\lim\limits_{\rho \rightarrow 0^{+}} \varphi(\rho) = 0$, such that
the following statement holds with high probability:
\begin{equation}
\|\mathcal{M}\| \leq \varphi({\rho}) \sqrt{n_{(1)}n_3}. \label{eq27}
\end{equation}
\label{lem4}
\end{lemma}

\subsection*{Subgradients of tensor $l_{1}$ norm and transformed tubal nuclear norm}
In the optimalization model (\ref{mainp}), tensor $l_{1}$ norm and
transformed tubal nuclear
 norm are used. One of the main technical tools in analyzing the tensor norms minimization is the characterization of the
 subgradients.  Then we list the subdifferential of
 tensor $l_{1}$ norm and a subset of the subdifferential of
transformed tubal nuclear
 norm, respectively.

Suppose that $\mathcal{A},\mathcal{ G} \in \C^{n_1\times n_2\times n_{3}}$, the
subdifferential of any norm $ \| |\mathcal{A} |\|$ can be given by
\begin{equation}\label{new6}
\begin{split}
\partial \| |\mathcal{A}|\|=\Big\{ \mathcal{ G}:\|| \mathcal{ B}|\| \geq \| |\mathcal{ A}| \| +& \textrm{Re} (\left\langle
\mathcal{ G}, \mathcal{ B}- \mathcal{ A}\right\rangle), \forall \ \mathcal{B}\in \C^{n_1\times n_2\times n_{3}}\Big\}.
\end{split}
\end{equation}
 Here $\textrm{Re}(\cdot)$ denotes the real part of a complex number.
 Let $\widetilde{\Omega}$ denotes the locations such that $\mathcal{A}_{ijk}$ are nonzeros.

\begin{lemma}\label{lem:4.8}
Suppose that $\mathcal{A},\mathcal{ G} \in \C^{n_1\times n_2\times n_{3}}$,
denote $\partial \| \mathcal{A} \|_{1}$ as the subdifferential of $\left \Vert \cdot \right \Vert _{1}$ at
$\mathcal{A}$ wihch is supported on $\widetilde{\Omega}$, then
%subgradient of $\left \Vert \cdot \right \Vert _{1}$ at
%$\mathcal{A}$ supported on $\Omega$ is given by
\begin{equation}
\partial \| \mathcal{A} \|_{1}=
\left \{ \mathcal{G}:
\mathcal{G}= {\tt dir}(\mathcal{A} )+\mathcal{F} ,
\mathcal{P}_{\widetilde{\Omega}}\left(\mathcal{F} \right) =0, \| \mathcal{F}
\|_{\infty}\leq1\right \},  \label{0.4}
\end{equation}
where ${\tt dir}(\mathcal{A} )$ is defined as
\begin{equation*}
{\tt
dir}(\mathcal{A}_{ijk}) = \left\{
\begin{array}{lll}
\frac{\mathcal{A}_{ijk}}{|\mathcal{A}_{ijk}|}, & \mathcal{A}_{ijk}\neq 0,\\
0, & \mathcal{A}_{ijk}=0.\\
\end{array}
\right.
\end{equation*}
\end{lemma}
\noindent
%{\bf Proof of Lemma \ref{lem:4.8}:}
{\bf Proof.}
For any $x\in\mathbb{C}$, the subdifferential of $|\cdot|$ is given by
\[
\partial |x| = \left\{
\begin{array}{lll}
x/|x|, & \mbox{if}  \ x\neq 0,\\
$[-1,1]$, & \mbox{if} \ x=0.\\
\end{array}
\right.
\]
Since the each entries of $\|\mathcal{A}\|_1$ is separable, we can obtain easily that
$$
\partial \| \mathcal{A} \|_{1}= \left \{ \mathcal{G}:
\mathcal{G}= {\tt dir}(\mathcal{A} )+\mathcal{F} ,
\mathcal{P}_{\widetilde{\Omega}}\left(\mathcal{F} \right) =\mtx{0}, \| \mathcal{F}
\|_{\infty}\leq1\right \}.
$$
This completes the proof. \qed

\begin{lemma}\label{lem:4.9}
Suppose that $\mathcal{A}\in \C^{n_1\times n_2\times n_{3}} $ has
the tensor singular value decomposition as \eqref{svd1}. Denote
$\partial \| \mathcal{A} \|_{\textup{TTNN}}$  as the subdifferential of $\left
\Vert \cdot \right \Vert _{\textup{TTNN}}$ at $\mathcal{A},$ then
\begin{equation}\label{0.2}
\partial \| \mathcal{A} \|_{\textup{TTNN}}\supseteq
\left \{ \mathcal{G} \in \C^{n_1\times n_2\times n_{3}}:
\mathcal{G}= \mathcal{U} \diamond_{\bf \Phi}
\mathcal{V}^{H}+\mathcal{F}\right\},
\end{equation}
where $~ \mathcal{U}^{H}\diamond_{\bf
\Phi}\mathcal{F}=\mathcal{F}^{H} \diamond_{\bf \Phi} \mathcal{V}=\mtx{0}~ \text{and}~
\| \mathcal{F} \| \le 1.$
\end{lemma}

%\begin{remark} For simplicity, we only derive the expressions of two subsets  of the subgradients relate to tensor $l_{1}$ norm and transformed nuclear norm, which are used in the dual  proof.
%The necessary conditions can be derived
%by the results in \cite{KZ,GAW}.
%\end{remark}

\noindent
%{\bf Proof of Lemma \ref{lem:4.9}:}
{\bf Proof.}
If $\mathcal{A}=\mtx{0}$, then the result is obvious. We now assume
$\mathcal{A} \neq \mtx{0}$, and $\mathcal{Y}$ is given as (\ref{0.2}). It
follows that
\begin{align*}
 \textrm{Re}(\left\langle \mathcal{A},\mathcal{Y}\right\rangle) &
 =\textrm{Re}\left(\langle\mathcal{A},\mathcal{U} \diamond_{\bf \Phi}
\mathcal{V}^{H}+\mathcal{F}\rangle\right)=\textrm{Re}\left(\langle\mathcal{A},\mathcal{U}
\diamond_{\bf \Phi}
\mathcal{V}^{H}\rangle\right)+\textrm{Re}\left(\langle\mathcal{A},\mathcal{F}\rangle\right)
=\|\mathcal{A}\|_{\textup{TTNN}}.
\end{align*}
Moreover,
\begin{equation*}
\left\Vert \mathcal{Y}\right\Vert=\left\Vert\mathcal{U}
\diamond_{\bf \Phi} \mathcal{V}^{H} +\mathcal{F}\right\Vert =\max
\left\{ \left\Vert\mathcal{U} \diamond_{\bf \Phi}
\mathcal{V}^{H}\right \Vert +\left\Vert\mathcal{F}\right\Vert
\right\} =1.
\end{equation*}%
Then $\mathcal{Y}$  satisfies \eqref{new6} which is saying that $\mathcal{Y}$ a subgrandient of the tensor transform nuclear
norm at $\mathcal{A}$. \qed

The argument of the proof of Theorem \ref{TheoremM} can be divided into two
steps. The first step is to show a sufficient condition for the
pair $(\mathcal{L}_0, \mathcal{E}_0)$ to be the unique optimal
solution to problem (\ref{mainp}). The second step is to prove that
when required assumption in Theorem \ref{TheoremM} are satisfied, the
sufficient conditions derive by the first step is satisfied.  Now we
give a sufficient condition for the pair $(\mathcal{L}_0,
\mathcal{E}_0)$ to be the unique optimal solution to problem
(\ref{mainp}). The conditions are stated in terms of a dual variable
$\mathcal{Y}$, which is given in Theorem \ref{the3}.

\begin{theorem}\label{the3}
Assume that there is a tensor $\mathcal{Y} \in \C^{n_1 \times n_2 \times
n_3}$ obeying
\begin{equation}
\left\{
\begin{array}{lll}
\|\PT(\mathcal{Y} + \lambda{\tt dir}(\mathcal{E}_0) - \mathcal{U}\diamond_{\bf \Phi}\mathcal{V}^{H})\|_F \leq \frac{\lambda}{n_1n_2n^2_3}, \\
\|\PTc(\mathcal{Y} + \lambda{\tt dir}(\mathcal{E}_0))\| \leq \frac{1}{2}, \\
\|\Pg(\mathcal{Y})\|_{\infty} \leq \frac{\lambda}{2},\\
\Pgc(\mathcal{Y}) = \mtx{0},
\end{array}
\right. \label{eq28}
\end{equation}
where $\lambda = 1/\sqrt{\rho n_{(1)}n_3}$, then $(\mathcal{L}_0,
\mathcal{E}_0)$ is the unique optimal solution to (\ref{mainp}) when
$n_1, n_2, n_3$ are sufficient large.
\end{theorem}

\noindent
%{\bf Proof of Theorem 4:}
{\bf Proof.}
Let $f(\mathcal{L}, \mathcal{E}) := \|\mathcal{L}\|_{\textup{TTNN}} +
\lambda \|\mathcal{E}\|_1$. Given a feasible perturbation
$(\mathcal{L}_0 + \mathcal{Z}, \mathcal{E}_0 - \PO(\mathcal{Z}))$,
we will show that the objective function value $f(\mathcal{L}_0 +
\mathcal{Z}, \mathcal{E}_0 - \PO(\mathcal{Z}))$ is strictly larger
than $f(\mathcal{L}_0, \mathcal{E}_0)$ unless $\mathcal{Z} =
\mtx{0}$. Denote the transformed tensor SVD of $\PTc(\mathcal{Z})$ by
$\PTc(\mathcal{Z}) = \mathcal{U}_\perp \diamond_{\bf \Phi}\mathcal{S}_\perp \diamond_{\bf \Phi} \mathcal{V}^{H}_\perp$.
Since $\mathcal{\overline{U}}^{H}\mathcal{\overline{U}}_{\perp}=\mtx{0}$
and $\mathcal{\overline{V}}^{H}\mathcal{\overline{V}}_{\perp}=\mtx{0}$, we
have
$$
\|\mathcal{U} \diamond_{\bf \Phi} \mathcal{V}^{H} +
\mathcal{U}_\perp \diamond_{\bf \Phi} \mathcal{V}^{H}_\perp\| =
\|\mathcal{\overline{U}}~ \mathcal{\overline{V}}^{H} +
\mathcal{\overline{U}}_\perp \mathcal{\overline{V}}^{H}_\perp\|=1.
$$
It follows from Lemmas \ref{lem:4.8} and \ref{lem:4.9} that
\begin{align}
\|\mathcal{L}_0 + \mathcal{Z}\|_{\text{TTNN}} & \geq \textrm{Re}(
\langle\mathcal{U} \diamond_{\bf \Phi} \mathcal{V}^{H}
+ \mathcal{U}_\perp \diamond_{\bf \Phi} \mathcal{V}^{H}, \mathcal{L}_{0} + \mathcal{Z}\rangle )\nonumber \\
& = \textrm{Re}( \langle\mathcal{U} \diamond_{\bf \Phi}
\mathcal{V}^{H},\mathcal{L}_{0}\rangle+\langle\mathcal{U}_\perp \diamond_{\bf \Phi} \mathcal{V}^{H}_\perp,
\PTc(\mathcal{Z})\rangle +\langle\mathcal{U} \diamond_{\bf \Phi} \mathcal{V}^{H}, \mathcal{Z}\rangle)  \nonumber\\
& =  \|\mathcal{L}_0\|_{\textup{TTNN}} +
\|\PTc(\mathcal{Z})\|_{\textup{TTNN}} +\textrm{Re}(\langle
\mathcal{U} \diamond_{\bf \Phi} \mathcal{V}^{H},
\mathcal{Z}\rangle), \nonumber
\end{align}
and
\begin{align}
 \|\mathcal{E}_0 - \PO(\mathcal{Z})\|_1  &= \|\PO(\mathcal{E}_0 -\mathcal{Z})\|_1  = \|\Pg(\mathcal{E}_0 - \mathcal{Z})\|_1 + \|\Pl(\mathcal{E}_0 - \mathcal{Z})\|_1  \nonumber\\
& =  \|\Pg(\mathcal{Z})\|_1 + \|\mathcal{E}_0 -\Pl(\mathcal{Z})\|_1
 \geq  \|\Pg(\mathcal{Z})\|_1 + \|\mathcal{E}_0\|_1 -\textrm{Re}(\langle{\tt dir}(\mathcal{E}_0), \Pl(\mathcal{Z})\rangle)\nonumber\\
& \geq  \|\Pg(\mathcal{Z})\|_1 + \|\mathcal{E}_0\|_1
-\textrm{Re}(\langle{\tt dir}(\mathcal{E}_0), \mathcal{Z}\rangle).
\nonumber
\end{align}
Therefore, we obtain that
\begin{align}
&~~~~~f(\mathcal{L}_0 + \mathcal{Z}, \mathcal{E}_0 - \PO(\mathcal{Z})) - f(\mathcal{L}_0, \mathcal{E}_0) \nonumber \\
& = \|\mathcal{L}_0 + \mathcal{Z}\|_{\textup{TTNN}} + \lambda\|\mathcal{E}_0 - \PO(\mathcal{Z})\|_1 - \|\mathcal{L}_0\|_{\textup{TTNN}} - \lambda\|\mathcal{E}_0\|_1 \nonumber\\
& \geq  \|\PTc(\mathcal{Z})\|_{\textup{TTNN}} + \lambda
\|\Pg(\mathcal{Z})\|_1 -\textrm{ Re}\<\lambda {\tt
dir}(\mathcal{E}_0) - \mathcal{U}
\diamond_{\bf \Phi} \mathcal{V}^{H}, \mathcal{Z}\> \nonumber\\
&\geq \|\PTc(\mathcal{Z})\|_{\textup{TTNN}} + \lambda
\|\Pg(\mathcal{Z})\|_1 - \textrm{ Re}\<\Pg(\mathcal{Y}), \Pg(\mathcal{Z})\>\nonumber\\
&~~~~~-\textrm{ Re}\<\PT(\mathcal{Y} + \lambda {\tt
dir}(\mathcal{E}_0) - \mathcal{U}
\diamond_{\bf \Phi} \mathcal{V}^{H}), \PT(\mathcal{Z})\> -\textrm{ Re}\<\PTc(\mathcal{Y} + \lambda {\tt dir}(\mathcal{E}_0)), \PTc(\mathcal{Z})\>  \nonumber\\
& = \|\PTc(\mathcal{Z})\|_{\textup{TTNN}} + \lambda \|\Pg(\mathcal{Z})\|_1- \textrm{ Re}\<\Pg(\mathcal{Y}), \Pg(\mathcal{Z})\>\nonumber\\
&~~~~~-\textrm{ Re}\<\overline{\PT(\mathcal{Y} + \lambda {\tt dir}(\mathcal{E}_0) - \mathcal{U}\diamond_{\bf \Phi} \mathcal{V}^{H})}, \overline{\PT(\mathcal{Z})}\> -\textrm{ Re}\<\overline{\PTc(\mathcal{Y} + \lambda {\tt dir}(\mathcal{E}_0))}, \overline{\PTc(\mathcal{Z})}\>  \nonumber\\
& \geq  \|\PTc(\mathcal{Z})\|_{\textup{TTNN}} + \lambda \|\Pg(\mathcal{Z})\|_1- \|\Pg(\mathcal{Y})\|_{\infty}\|\Pg(\mathcal{Z})\|_1 \nonumber\\
&~~~~~-\|\overline{\PT(\mathcal{Y} + \lambda {\tt dir}(\mathcal{E}_0) - \mathcal{U} \diamond_{\bf \Phi} \mathcal{V}^{H})}\|_F \|\overline{\PT(\mathcal{Z})}\|_F -\|\overline{\PTc(\mathcal{Y} + \lambda {\tt dir}(\mathcal{E}_0))}\| \|\overline{\PTc(\mathcal{Z})}\|_{\ast} \nonumber\\
& =  \|\PTc(\mathcal{Z})\|_{\textup{TTNN}} + \lambda
\|\Pg(\mathcal{Z})\|_1- \|\Pg(\mathcal{Y})\|_{\infty}\|\Pg(\mathcal{Z})\|_1\nonumber\\
&~~~~~-\|\PT(\mathcal{Y} + \lambda {\tt
dir}(\mathcal{E}_0) - \mathcal{U}
\diamond_{\bf \Phi} \mathcal{V}^{H})\|_F \|\PT(\mathcal{Z})\|_F - \|\PTc(\mathcal{Y} + \lambda {\tt dir}(\mathcal{E}_0))\| \|\PTc(\mathcal{Z})\|_{\textup{TTNN}}  \nonumber\\
& \geq   \frac{1}{2}\|\PTc(\mathcal{Z})\|_{\textup{TTNN}} +
\frac{\lambda}{2}\|\Pg(\mathcal{Z})\|_1 -\frac{\lambda}{n_1n_2n^2_3} \|\PT(\mathcal{Z})\|_F, \label{eq29}
\end{align}
where the inequality (\ref{eq29}) is due to (\ref{eq28}).
It follows from Lemma \ref{le1} that
$$\Big\|\frac{1}{(1-2\gamma)\rho}\PT\Pg\PT - \PT\Big\|_{\textup{op}}
\leq \frac{1}{2},
$$ which implies
$\|\frac{1}{\sqrt{(1-2\gamma)\rho}}\PT\Pg\|_{\textup{op}} \leq
\sqrt{3/2}$. Therefore, we get
\begin{align}
&~~~~~\|\PT(\mathcal{Z})\|_F  = \|\overline{\PT(\mathcal{Z})}\|_F
\nonumber\\ &\leq 2 \Big\|\frac{1}{(1-2\gamma)\rho}\overline{\PT\Pg\PT(\mathcal{Z})}\Big\|_F \leq  2 \Big\|\frac{1}{(1-2\gamma)\rho}\overline{\PT\Pg\PTc(\mathcal{Z})}\Big\|_F
+ 2 \Big\|\frac{1}{(1-2\gamma)\rho}\overline{\PT\Pg(\mathcal{Z})}\Big\|_F \nonumber\\
%& \leq  \sqrt{\frac{6}{(1-2\gamma)\rho }}\|\overline{\PTc(\mathcal{Z})}\|_F + \sqrt{\frac{6}{(1-2\gamma)\rho }}\|\overline{\Pg(\mathcal{Z})}\|_F \nonumber\\
& \leq  \sqrt{\frac{6}{(1-2\gamma)\rho}}\|\PTc(\mathcal{Z})\|_F +
\sqrt{\frac{6}{(1-2\gamma)\rho}}\|\Pg(\mathcal{Z})\|_F.\label{eq30}
\end{align}
It is easy to check that
\begin{equation}
\|\PTc(\mathcal{Z})\|_{\textup{TTNN}} =
\|\overline{\PTc(\mathcal{Z})}\|_{\ast} \geq
\|\overline{\PTc(\mathcal{Z})}\|_F = \|\PTc(\mathcal{Z})\|_F
\label{eq31}
\end{equation}
and $\|\Pg(\mathcal{Z})\|_1 \geq \|\Pg(\mathcal{Z})\|_F$.
Substituting (\ref{eq30}) and (\ref{eq31}) into (\ref{eq29}), we
have
\begin{align}
&f(\mathcal{L}_0 + \mathcal{Z}, \mathcal{E}_0 - \PO(\mathcal{Z})) - f(\mathcal{L}_0, \mathcal{E}_0) \nonumber\\
\geq &\Bigg(\frac{1}{2} -
\frac{\lambda}{n_1n_2n^2_3}\sqrt{\frac{6}{(1-2\gamma)\rho}}\Bigg)\|\PTc(\mathcal{Z})\|_F + \Bigg(\frac{\lambda}{2} -
\frac{\lambda}{n_1n_2n^2_3}\sqrt{\frac{6}{(1-2\gamma)\rho}}\Bigg)\|\Pg(\mathcal{Z})\|_F.
\label{eq32}
\end{align}
When $n_1, n_2, n_3$ are sufficiently large such that
\begin{equation}
\frac{1}{2} - \frac{\lambda}{n_1 n_2
n^2_3}\sqrt{\frac{6}{(1-2\gamma)\rho}} > 0, ~ \frac{\lambda}{2}
- \frac{\lambda}{n_1 n_2 n^2_3}\sqrt{\frac{6}{(1-2\gamma)\rho}} > 0,
\nonumber
\end{equation}
the inequality (\ref{eq32}) holds if and only if $\PTc(\mathcal{Z})
= \Pg(\mathcal{Z}) = \mtx{0}$. Thus, $ \PT\Pg\mathcal{(Z)}=\mtx{0}.$
 On the other hand, when $\rho$ is sufficiently large and $\gamma$ is
sufficiently small (which are bounded by two constants $c_\rho$ and
$c_\gamma$), we obtain that
\begin{equation}
\|\PT\Pg\|_{\textup{op}} \leq \sqrt{\frac{3(1 - 2\gamma)\rho}{2}} <
1, \nonumber
\end{equation}
which implies that $\PT\Pg$ is injective. As a result, (\ref{eq32})
holds if and only if $\mathcal{Z} = \mtx{0}$. \qed

Similar to \cite{Candes2011} and \cite{Gross2010}, we use the golfing scheme to construct the dual tensor
$\mathcal{Y}$, which is supported on $\Gamma$. Afterwards, the size of $\Gamma$ is increased gradually.
Consider the set $\Gamma \sim
\textup{Ber}((1-2\gamma)\rho)$ as a union of sets of support
$\Gamma_j$, i.e.,
$$
\Gamma = \bigcup_{j=1}^p \Gamma_j
$$
where
$\Gamma_j \sim \textup{Ber}(q_j)$. Let $q_1 = q_2 =
\frac{(1-2\gamma)\rho}{6}$ and $q_3 = \dots = q_p = q$, which
implies $q \geq C_0\rho/\log(n_{(1)}n_3)$. Hence we have
\begin{equation*}
1 - (1 - 2\gamma)\rho = \Big(1 - \frac{(1-2\gamma)\rho}{6}\Big)^2 (1
- q)^{p-2}, \label{eq33}
\end{equation*}
where $p = \lfloor 5\log(n_{(1)}n_3) + 1\rfloor$. Starting from
$\mathcal{Z}_0 = \PT(\mathcal{U} \diamond_{\bf \Phi} \mathcal{V}^{H}
- \lambda \sgn(\mathcal{E}_0))$, we define inductively
\begin{equation*}
\mathcal{Z}_{j} = \Big(\PT -
\frac{1}{q_j}\PT\Pgs{j}\PT\Big)\mathcal{Z}_{j-1}. \label{eq34}
\end{equation*}
Then it follows from Lemmas \ref{le1}-\ref{lem3} that
\begin{align}
\|\mathcal{Z}_j\|_F &\leq \frac{1}{2}\|\mathcal{Z}_{j-1}\|_F, \,\,\,j
= 1, \dots,p, \label{eq35}\\
\|\mathcal{Z}_1\|_{\infty} &\leq \frac{1}{2\sqrt{\log(n_{(1)}n_3)}}\|\mathcal{Z}_0\|_{\infty},\label{eq36}\\
\|\mathcal{Z}_j\|_{\infty} &\leq
\frac{1}{2^j\log(n_{(1)}n_3)}\|\mathcal{Z}_0\|_{\infty}, \,\,\,j =
2, \dots, p,\label{eq37}
\end{align}
and
\begin{align}
\|(\OpId_{\bf \Phi} - q^{-1}_j\Pgs{j}) \mathcal{Z}_{j-1}\| \leq C'_0
\sqrt{\frac{n_{(1)}n_3\log(n_{(1)}n_3)}{q_j}}\|\mathcal{Z}_{j-1}\|_{\infty},\,\,\,j
= 2, \dots, p, \label{eq38}
\end{align}
with high probability provided $c_r$ and $c_\gamma$ are small
enough.

Let the dual tensor $\mathcal{Y}$ be
\begin{equation}
\mathcal{Y} = \sum_{j=1}^p \frac{1}{q_j}\Pgs{j}(\mathcal{Z}_{j-1}).
\label{eq39}
\end{equation}
Then we need to show that the $\mathcal{Y}$ in (\ref{eq39}) satisfies (\ref{eq28}). Obviously,
$\Pgc(\mathcal{Y}) = \mtx{0}$ and it is sufficient to prove
\begin{equation}
\left\{
\begin{array}{llll}
\|\PT(\mathcal{Y} + \lambda{\tt dir}(\mathcal{E}_0) - \mathcal{U}
\diamond_{\bf \Phi}
\mathcal{V}^{H})\|_F \leq \frac{\lambda}{n_1n_2n^2_3}, \\
\|\PTc(\mathcal{Y})\| \leq \frac{1}{4}, \\
\lambda\|\PTc({\tt dir}(\mathcal{E}_0))\| \leq \frac{1}{4}, \\
\|\Pg(\mathcal{Y})\|_{\infty} \leq \frac{\lambda}{2},
\end{array}
\right. \label{eq40}
\end{equation}
where $\lambda = 1/\sqrt{\rho n_{(1)}n_3}$ and $n_1, n_2, n_3$ are
large enough.

First, let us bound $\|\mathcal{Z}_0\|_F$ and
$\|\mathcal{Z}_0\|_{\infty}$. By the triangle inequality, we get
\begin{equation}\label{zif}
\|\mathcal{Z}_0\|_{\infty} \leq \|\mathcal{U} \diamond_{\bf \Phi}
\mathcal{V}^{H}\|_{\infty} +
\lambda\|\PT({\tt
dir}(\mathcal{E}_0))\|_{\infty}.
\end{equation}
Note that $\mathcal{E}_{ijk} ={\bf \Phi}[ \tc{e}_i\diamond_{\bf \Phi}
\tub{e}_k \diamond_{\bf \Phi} \tc{e}_j]$ and denote $(\mathcal{E}_{\bf \Phi})_{ijk}=\tc{e}_i\diamond_{\bf \Phi}
\tub{e}_k \diamond_{\bf \Phi} \tc{e}_j,$  then
\begin{align*}
{\tt dir}(\mathcal{E}_0) &= \sum_{i,j,k} \big({\tt
dir}(\mathcal{E}_0)\big)_{i,j,k} \Phi[\tc{e}_i\diamond_{\bf
\Phi}\tub{e}_k\diamond_{\bf \Phi}\tc{e}_j^{H}]=\sum_{i,j,k} \big({\tt
dir}(\mathcal{E}_0)\big)_{i,j,k} \Phi[(\mathcal{E}_{\bf \Phi})_{ijk}].
\end{align*}
Therefore, by (\ref{j01}) we have
\begin{align*}
\PT({\tt dir}(\mathcal{E}_0)) &=\sum_{i,j,k}\big({\tt
dir}(\mathcal{E}_0)\big)_{i,j,k}\PT(\Phi[\tc{e}_i\diamond_{\bf
\Phi}\tub{e}_k\diamond_{\bf \Phi}\tc{e}_j^{H}])\\
 &=
\sum_{i,j,k}\big({\tt
dir}(\mathcal{E}_0)\big)_{i,j,k}\left\langle\mathcal{E}_{ijk},\PT((\mathcal{E}_{\bf \Phi})_{ijk})\right\rangle
(\mathcal{E}_{\bf \Phi})_{ijk}.
\end{align*}
Hence, the $(a,b,c)$-th entry of $\PT({\tt dir}(\mathcal{E}_0))$ can be
represented by
\begin{align*}
&~~~\<\PT({\tt dir}(\mathcal{E}_0)), \Phi[\tc{e}_a\diamond_{\bf
\Phi}\tub{e}_b\diamond_{\bf \Phi}\tc{e}_c^{H}]\> = \sum_{ijk}
\big({\tt
dir}(\mathcal{E}_0)\big)_{i,j,k}\<\PT((\mathcal{E}_{\bf \Phi})_{ijk}),
\Phi[(\mathcal{E}_{\bf \Phi})_{abc}]\>.
\end{align*}
By Bernstein's inequality, we further have
\begin{align*}
\PP(|\<\PT({\tt dir}(\mathcal{E}_0)), \Phi[(\mathcal{E}_{\bf \Phi})_{ijk}]\>| \geq \tau)
 \leq
2\exp\bigg(\frac{-\tau^2/2}{N + M\tau/3}\bigg), \nonumber
\end{align*}
where
\begin{align*}
 M\nonumber&= \big|\big[{\tt dir}(\mathcal{E}_0)\big]_{i,j,k} \ \big|  \
\|\PT((\mathcal{E}_{\bf \Phi})_{ijk})\|_F \|\PT((\mathcal{E}_{\bf \Phi})_{abc})\|_F \leq \frac{2\mu
r}{n_{(2)}},
\end{align*}
and $N = 2\gamma\rho \|\PT((\mathcal{E}_{\bf \Phi})_{ijk})\|^2_F \leq
4\gamma\rho\frac{\mu r}{n_{(2)}}.$
Since the entries of $\PT({\tt dir}(\mathcal{E}_0))$ can be
understood as i.i.d. copies of the $(a, b, c)$-th entry, it follows from
the union bound that
\begin{equation}\label{PIZ}
\|\PT({\tt dir}(\mathcal{E}_0))\|_{\infty} \leq C''
\sqrt{\frac{\rho \mu r \log(n_{(1)}n_3)}{n_{(2)}}}
\end{equation}
with high probability for some numerical constant $C''$.
By the joint incoherence condition (\ref{eq18}), we obtain that
\begin{equation}\label{UVI}
\|\mathcal{U}\diamond_{\bf \Phi} \mathcal{V}^{H}\|_{\infty} \leq
\sqrt{\frac{\mu r}{n_1 n_2 n_3}} = \lambda \sqrt{\frac{\rho \mu
r}{n_{(2)}}}.
\end{equation}
Therefore, combining (\ref{zif}), (\ref{PIZ}), and (\ref{UVI}), we get
\begin{align}
\|\mathcal{Z}_0\|_{\infty} & \leq C \lambda \sqrt{\frac{\rho \mu r\log(n_{(1)}n_3)}{n_{(2)}}}, \label{eq41}\\
\|\mathcal{Z}_0\|_{F} & \leq
\sqrt{n_1n_2n_3}\|\mathcal{Z}_0\|_{\infty} \leq C \lambda \sqrt{\rho
\mu rn_{(1)}n_3\log(n_{(1)}n_3)}, \label{eq42}
\end{align}
where $C = \max\big\{\frac{1}{\log(n_{(1)}n_3)}, C''\big\}$.

Now, let us turn to the proof of (\ref{eq40}).
By (\ref{eq39}), we obtain that
\begin{align}
&\|\PT(\mathcal{Y}) + \PT(\lambda{\tt dir}(\mathcal{E}_0) -
\mathcal{U} \diamond_{\bf \Phi} \mathcal{V}^{H})\|_F
=\Big\|\mathcal{Z}_0 -\sum_{j=1}^p \frac{1}{q_j}\PT\Pgs{j}(\mathcal{Z}_{j-1})\Big\|_F\nonumber\\
=~&  \Big\|(\PT - \frac{1}{q_1}\PT\Pgs{1}\PT)\mathcal{Z}_0 -
\sum_{j=2}^p \frac{1}{q_j}\PT\Pgs{j}\PT(\mathcal{Z}_{j-1})
\Big\|_F \nonumber\\\
= ~& \Big\|\PT(\mathcal{Z}_1)-\sum_{j=2}^p
\frac{1}{q_j}\PT\Pgs{j}\PT(\mathcal{Z}_{j-1})\Big\|_F
= ~\dots = \|\mathcal{Z}_p\|_F\nonumber\\
\leq ~&\Big(\frac{1}{2}\Big)^p \|\mathcal{Z}_0\|_F \leq  C \Big(n_{(1)}n_3\Big)^{-5} \lambda \sqrt{\rho \mu r n_{(1)}
n_3 \log(n_{(1)}n_3)}\nonumber\\
\leq ~& \frac{\lambda}{n_1n_2n^2_3}, \nonumber
\end{align}
where the first inequality follows from (\ref{eq35}) and the second inequality from
(\ref{eq42}).

Furthermore, provided that $\lambda = 1/\sqrt{\rho n_{(1)}n_3}$ and $c_r$ is
sufficiently small, we have
\begin{align}
&\|\PTc(\mathcal{Y})\|
 = \Big\|\PTc \sum_{j=1}^p
\frac{1}{q_j}\Pgs{j}(\mathcal{Z}_{j-1})\Big\|
\nonumber\\ \leq ~&\sum_{j=1}^p \Big\|\frac{1}{q_j} \PTc\Pgs{j}(\mathcal{Z}_{j-1})\Big\|
 =\sum_{j=1}^p
\Big\|\PTc\Big(\frac{1}{q_j}\Pgs{j}(\mathcal{Z}_{j-1}) -
\mathcal{Z}_{j-1}\Big)\Big\| \nonumber\\
 \leq~&  \sum_{j=1}^p \Big\|\frac{1}{q_j}\Pgs{j}(\mathcal{Z}_{j-1}) - \mathcal{Z}_{j-1}\Big\|
 \leq\sum_{j=1}^p C'_0 \sqrt{\frac{n_{(1)}n_3\log(n_{(1)}n_3)}{q_j}}\|\mathcal{Z}_{j-1}\|_{\infty}  \nonumber\\
\leq~&  C'_0 \sqrt{n_{(1)}n_3\log(n_{(1)}n_3)}\Big(\sum_{j=3}^p
\frac{1}{2^{j-1}\log(n_{(1)}n_3)\sqrt{q_j}}
 +
 \frac{1}{2\sqrt{\log(n_{(1)}n_3)}\sqrt{q_2}} +\frac{1}{\sqrt{q_1}}\Big)\|\mathcal{Z}_0\|_{\infty} \nonumber\\
 \leq ~&C' \lambda \sqrt{\frac{ \rho \mu r n_{(1)}n_3(\log(n_{(1)}n_3))^2}{\rho n_{(2)}}}\nonumber\\
\leq~&  C' \sqrt{c_r} \leq \frac{1}{4}, \nonumber
\end{align}
where the third inequality follows  from Lemma~\ref{lem3}, the
fourth inequality follows from (\ref{eq36})-(\ref{eq38}), and the fifth inequality follows from
(\ref{eq41}), respectively.

Note that the direction tensor ${\tt dir}(\mathcal{E}_0)$ satisfies
\begin{equation}
\big|\big[{\tt dir}(\mathcal{E}_0)\big]_{i,j,k}\big| = \left\{
\begin{array}{lll}
1, & \textup{with \ probability}\,\,\gamma\rho,\\
0, & \textup{with \ probability}\,\,1-\gamma\rho.\\
\end{array}
\right. \nonumber
\end{equation}
As proved by Lemma~\ref{lem4}, there exists a function
$\varphi(\gamma\rho)$ satisfying $\lim\limits_{\gamma\rho
\rightarrow 0^{+}} \varphi(\gamma\rho) = 0$, such that
\begin{equation}
\|{\tt dir}(\mathcal{E}_0)\| \leq \varphi(\gamma\rho)
\sqrt{n_{(1)}n_3} \nonumber
\end{equation}
with high probability, which yields
\begin{equation*}
\lambda\|\PTc({\tt dir}(\mathcal{E}_0))\| \leq \lambda\|{\tt
dir}(\mathcal{E}_0)\| \leq \varphi(\gamma\rho)/\sqrt{\rho} \leq
\frac{1}{4},
\end{equation*}
as long as $c_r$ and $c_\gamma$ is sufficiently small.

Finally, when $c_r$ is small enough, we can get
\begin{align}
\|\Pg(\mathcal{Y})\|_{\infty}
&= \Big\|\PT\sum_{j=1}^p
\frac{1}{q_j}\Pgs{j}(\mathcal{Z}_{j-1})\Big\|_{\infty} \leq \sum_{j=1}^p \frac{1}{q_j}\|\mathcal{Z}_{j-1}\|_{\infty}\nonumber\\
& \leq \Big(\sum_{j=3}^p \frac{1}{2^{j-1}\log(n_{(1)}n_3)\sqrt{q_j}}\Big)\|\mathcal{Z}_0\|_{\infty}+\Big( \frac{1}{2\sqrt{\log(n_{(1)}n_3)q_2}} +\frac{1}{\sqrt{q_1}}\Big)\|\mathcal{Z}_0\|_{\infty}\nonumber\\
& \leq C \lambda \sqrt{\frac{\mu r\log(n_{(1)}n_3)}{\rho n_{(2)}}}\leq C\lambda \sqrt{\frac{c_r}{\log(n_{(1)}n_3)}} \leq
\frac{\lambda}{2}, \nonumber
\end{align}
where the first inequality follows from
Lemma~\ref{lem2}, the second inequality follows from \eqref{eq36}-\eqref{eq38}, and
the third inequality follows from \eqref{eq41}, respectively.

\bibliographystyle{abbrv}

\bibliography{RefRTC}

\end{document}